\documentclass[twoside,11pt]{article}

\usepackage{blindtext}

\usepackage[preprint,abbrvbib]{jmlr2e}

\hypersetup{hidelinks}

\usepackage{amsmath,bbold,bm,braket}
\usepackage[dvipsnames]{xcolor}
\usepackage{enumitem}

\newcommand{\bx}{\mathbf{x}}

\newcommand{\bX}{X}

\newcommand{\by}{\mathbf{y}}
\newcommand{\bs}{\mathbf{s}}
\newcommand{\bS}{\mathbf{S}}
\newcommand{\bZ}{\mathbf{Z}}
\newcommand{\CS}{\mathcal{S}}
\newcommand{\CN}{\mathcal{N}}
\newcommand{\CW}{\mathcal{W}}
\newcommand{\CL}{\mathcal{L}}
\newcommand{\CMN}{\mathcal{MN}}
\newcommand{\E}{\mathbf{E}}

\newcommand{\eI}{\mathbb{1}}
\newcommand{\FCK}{\mathcal{K}_{\mathrm{FC}}}
\newcommand{\CNN}{\mathcal{K}_{\mathrm{C}}}
\newcommand{\RE}{\mathbb{R}}

\DeclareMathOperator{\tr}{tr}
\DeclareMathOperator{\vvec}{vec}
\DeclareMathOperator{\GGamma}{Gamma}
\DeclareMathOperator{\Sp}{Sp}
\DeclareMathOperator{\Cov}{Cov}

\usepackage{lastpage}
\jmlrheading{26}{2025}{1-\pageref{LastPage}}{7/24; Revised
3/25}{4/25}{24-1158}{Federico Bassetti, Marco Gherardi, Alessandro Ingrosso, Mauro Pastore, and Pietro Rotondo}

\ShortHeadings{Feature learning in finite-width Bayesian deep linear networks}{Bassetti, Gherardi, Ingrosso, Pastore, and Rotondo}
\firstpageno{1}

\begin{document}

\title{Feature learning in finite-width Bayesian deep linear networks with multiple outputs and convolutional layers}

\author{\name Federico Bassetti 
       \\
       \addr Dipartimento di Matematica\\
       Politecnico di Milano\\
       Piazza Leonardo da Vinci 31, 20133 Milan, Italy
       \AND
       \name Marco Gherardi 
       \\
       \addr Dipartimento di Fisica\\
       Università degli Studi di Milano\\
       Via Celoria 16, 20133 Milan, Italy;\\
       Istituto Nazionale di Fisica Nucleare -- Sezione di Milano\\
       Via Celoria 16, 20133 Milan, Italy
       \AND
       \name Alessandro Ingrosso 
       \\
       \addr Donders Institute for Brain, Cognition and Behaviour\\
       Radboud University\\
       Nijmegen, The Netherlands
       \AND
       \name Mauro Pastore 
       \\
       \addr Laboratoire de Physique\\
       \'{E}cole Normale Sup\'{e}rieure, CNRS, PSL University, Sorbonne University, Universit\'{e} Paris-Cit\'{e}\\
       24 rue Lhomond, 75005 Paris, France;\\
       \addr Quantitative Life Sciences\\
       The Abdus Salam International Centre for Theoretical Physics\\
       Strada Costiera 11, 34151 Trieste, Italy
       \AND
       \name Pietro Rotondo 
       \\
       \addr Dipartimento di Scienze Matematiche, Fisiche e Informatiche\\
       Università degli Studi di Parma\\
       Parco Area delle Scienze 7/A, 43124 Parma, Italy
       }

\editor{Maxim Raginsky}

\maketitle

\begin{abstract}
Deep linear networks have been extensively studied, as they provide simplified models of deep learning. However, little is known in the case of finite-width architectures with multiple outputs and convolutional layers.
In this manuscript, we provide rigorous results for the statistics of functions implemented by the aforementioned class of networks, thus moving closer to a complete characterization of feature learning in the Bayesian setting. 
Our results include: (i) an exact and elementary non-asymptotic integral representation for the joint prior distribution over the outputs, given in terms of a mixture of Gaussians; (ii) an analytical formula for the posterior distribution in the case of squared error loss function (Gaussian likelihood); (iii) a quantitative description of the feature learning infinite-width regime, using large deviation theory.
From a physical perspective, deep architectures with multiple outputs or convolutional layers represent different manifestations of kernel shape renormalization, and our work provides a dictionary that translates this physics intuition and terminology into rigorous Bayesian statistics.
\end{abstract}


\section{Introduction}

Deep learning makes use of a wide array of architectures, each with unique strengths and applications. Selecting the optimal architecture for a specific task requires a deep understanding of their properties and a principled way to compute them from theory.
Unfortunately, the empirical advancements in this field risk outpacing 
the development of a comprehensive theoretical framework.

Significant progress has been made in certain asymptotic regimes, where neural networks enjoy forms of universality that simplify their analysis.
In the (lazy-training) infinite-width limit, in particular, 
universality is manifested in both the training under gradient flow, described by the neural tangent kernel (\citealp{JacotNTK}), and the Bayesian inference setting (\citealp{lee2018deep}), where exact relations between neural networks and kernel methods were obtained. 
However, these results cannot explain the success of modern deep architectures, which perform non-trivial feature selection beyond what is possible in the lazy-training regime (\citealp{ChizatLazy, lewkowycz2021the, NEURIPS2020_1457c0d6}). 
One can circumvent this problem by considering different rescalings of the weights with the width of the layers, using the so-called mean field (\citealp{doi:10.1073/pnas.1806579115,https://doi.org/10.1002/cpa.22074,doi:10.1137/18M1192184,NEURIPS2018_a1afc58c}) or the more recent maximal update parametrizations~(\citealp{yang2020feature}).

Despite these insights, neural networks at finite width, which are relevant for applications, display complex and intriguing phenomena that are not captured by infinite-width asymptotics (\citealp{stragglers, doi:10.1073/pnas.2201854119, Petrini_2023, DBLP:journals/corr/abs-2307-02129,doi:10.1073/pnas.2316301121, pmlr-v202-sclocchi23a}).
Addressing these finite-size properties calls for a non-perturbative theory, one which does not rely on large-size limits. Developing such a theory for general neural networks has proven to be quite challenging. 
However, progress has been made by considering simplified models where the activation functions are linear. Deep linear networks (\citealp{Saxelinear, doi:10.1073/pnas.1820226116, SompolinskyLinear}) are linear in terms of their inputs but they retain non-linear dependence on the parameters.
They provide analytically tractable non-convex problems in parameter space,
thus serving as a bridge between idealized infinite-width models and the reality of finite-sized networks.

Here, we address the computation of network statistics in the Bayesian learning setting, equivalent to the statistical mechanics framework employed by physicists
(\citealp{Neal, lee2018deep, novak2019bayesian, garriga-alonso2018deep,EngelVanDenBroeck}).
In Bayesian deep learning, the prior over the network's parameters induces a prior over the network's outputs
(more details are in Sec.~\ref{Sec:BayesianStting}).
In the (lazy-training) infinite-width limit, and for non-linear activation function, the prior over the outputs is known to be Gaussian both for 
fully connected architectures (\citealp{Neal, g.2018gaussian,Hanin2023})
and for convolutional architectures (\citealp{novak2019bayesian, garriga-alonso2018deep}).
For fully connected 
networks, precise rates of convergence to normality have been investigated as well (see  
\citealp{favaro2023quantitative,Trevisan2023}).
The goal of our work is to characterize the non-Gaussian behavior, at finite depth and width, of two classes of deep linear networks:
(i) those with fully-connected layers and multiple outputs, and
(ii) those with convolutional layers and a single linear readout.
Our results, summarized in the next section, provide a way to quantify non-perturbative feature learning effects in these architectures.
We will discuss the link between our work and the literature by physicists in Sec.~\ref{sec3}.

\subsection{Informal statement of the results}

We summarize here our main results informally, as three take-home messages.

\vspace{0.2cm}
\emph{Take-home message 1: 
 At finite width, the output prior is an exactly computable mixture of Gaussians. The sizes of the hidden layers appear as parameters in the mixing measure, leading to dimensional reduction.}

We  compute  the  prior over the outputs at finite-width in the linear case, showing that it is a  mixture of Gaussians with an explicit mixing distribution. 
Notably, in this representation the covariance, i.e. the neural network Gaussian process (NNGP) kernel, is
modified by $L$  (the number of hidden layers) random matrices with Wishart distribution.
For fully-connected linear networks with finite number $D$ of outputs in the readout layer, the dimension of these matrices is $D$ (\textbf{Proposition~\ref{prop1}}), while for convolutional linear networks (with unitary stride), the dimension of the Wishart matrices is the (fixed) size of the input $N_0$ (\textbf{Proposition~\ref{prop1-CNN}}).
All the sizes of the hidden layers $N_\ell$ appear parametrically in the prior, providing an explicit dimensional reduction.

\vspace{0.2cm}
\emph{Take-home message 2: At finite width
the posterior predictive is  a 
 mixture of Gaussians with closed form    
 mixing distribution.}
 
In the case of quadratic loss function (Gaussian likelihood), the posterior distribution 
inherits the properties of the prior. 
The posterior is again a mixture of Gaussians and this leads to the rather standard equations for the bias and variance of a Gaussian Process (\citealp{rasmussen2006}), with the important difference that they are now 
random variables {(\textbf{Proposition~\ref{Prop2} and~\ref{Prop2-CNN})}}.

\vspace{0.2cm}
\emph{Take-home message 3: 
In the feature learning infinite-width limit,
large deviation asymptotics shows non-trivial  
explicit dependence on the training inputs and labels.}

Using the simple parametric dependence of our formulas from the layer widths, 
we provide an asymptotic analysis in the limit of large width, using  the language  
of Large Deviation Theory. First, we recover the well-known 
infinite-width limit, showing that the prior is degenerate in this case, since the Wishart ensembles concentrate around the identity matrix (\textbf{Proposition~\ref{LDPeasy}}). 

Second, we consider the so-called feature learning infinite-width limit (\citealp{ChizatLazy, doi:10.1073/pnas.1806579115, GEIGER20211, Geiger_2020, yang2020feature}), which is equivalent to a different re-parametrization of the loss and of the output function, and we precisely show how it provides a way to escape lazy training. 
 This so-called mean field parametrization was initially investigated in deep networks trained using gradient descent (\citealp{NEURIPS2022_d027a5c9}), but it has been very recently considered also in the Bayesian setting (\citealp{rubin2024a,lauditi2025adaptive}).

The measure over the Wishart ensembles concentrates also in this case, but this time they do so around non-trivial solutions that explicitly depend on the training inputs and labels (\textbf{Proposition~\ref{LDP_postVara}}). 

\subsection{Related work\label{sec:related}}

\cite{Saxelinear, doi:10.1073/pnas.1820226116} derived exact solutions for the training dynamics of deep linear networks (DLN). More recently, \cite{SompolinskyLinear} studied deep linear networks in a Bayesian framework, via the equivalent statistical mechanics formulation, and obtained analytical results in the proportional limit, where the common size of the hidden layers $N$ and the size of the training set $P$ are taken to infinity, while keeping their ratio $\alpha = P/N$ fixed. This analysis was later generalized to globally gated DLNs (\citealp{li2022globally}), and extended to the data-averaged case, using the replica method from spin glass theory (\citealp{PhysRevE.105.064118}).

\cite{doi:10.1073/pnas.2301345120} reconsidered the same setting in the case of single-output networks, and they derived a non-asymptotic result for the Bayesian model evidence (partition function in statistical mechanics), in terms of Meijer G-functions (the fact that the prior over a single output is related to these special functions was shown in \citealp{zavatone-veth2021exact}). They employed this result to investigate various asymptotic limits where $N$, $P$, and the depth of the network $L$ are simultaneously taken to infinity at the same rate.

We provide an explicit mapping between our present results on one side, and both the statistical mechanics approach of~\cite{SompolinskyLinear} and the Meijer G-functions formalism of~\cite{doi:10.1073/pnas.2301345120,zavatone-veth2021exact} on the other side, in Section~\ref{sec:discussion}.

Closely related to the deep linear literature, \cite{pacelli2023statistical} proposed an effective theory for deep non-linear networks in the proportional limit, whose derivation is based on a Gaussian equivalence (\citealp{goldt2019hidden,goldt2020gaussian,pmlr-v119-20a, hu2020RF,gerace2022gaussian,aguirre2024random}) heuristically justified using Breuer-Major theorems (\citealp{BM, bardet2013, NourdinQuantitative}). This effective theory has been tested with great accuracy for one-hidden-layer fully-connected networks with a single output in~\citealp{baglioni2024predictive}. 
\citealp{camilli2023fundamental} and \citealp{cui2023optimal} reconsidered the same setting in the Bayes optimal and data-averaged case.
Other non-perturbative approaches to Bayesian deep non-linear networks include~\cite{NEURIPS2021_b24d2101,seroussi2023natcomm, FischerLindner:2024} and the recent renormalization group framework proposed in~\cite{howard2024wilsonian, howard2024bayesian}.

A further line of research considers  linear (or weakly nonlinear) networks in the limit in which both depth $N$ and width 
$L$ diverge, while their ratio $L/N$  converges to a positive constant; see \cite{hanin2024} and \cite{Mufan2023}. In particular, \cite{Mufan2023} finds that the conditional covariance matrix in the infinite-depth-and-width limit satisfies a stochastic differential equation for covariance matrices of dimension $P \times P$. 
Since the submission of the first version of this paper, further progress has been made in this direction. Building on our representation, a complete characterization of the proportional limit 
$L/N \to \alpha>0$  for deep linear networks is now available in the recent work \cite{BaLaRo}. This characterization has been made possible because our representation provides a fundamental dimensional reduction, which is not captured by the usual representation obtained by conditioning on the penultimate layer.

\subsection{Kernel renormalization in the proportional limit}

Bayesian networks in the infinite-width limit ($N\to\infty$ at finite $P$) are equivalent to Gaussian processes (NNGP),
whose covariance is given by a kernel (\citealp{lee2018deep}).
A powerful view of Bayesian networks in the proportional limit (\citealp{SompolinskyLinear, pacelli2023statistical, li2022globally}) focuses on how the prior is modified layer by layer.
For fully-connected architectures with a single output, the NNGP kernel at each layer is rescaled
by a real number (a scalar order parameter in statistical mechanics).
The order parameters satisfy a set of coupled equations that explicitly depend on the inputs and labels in the training set.
This process is referred to as a scalar renormalization of the kernel.
In fully-connected networks with multiple outputs, instead, the order parameters are $D \times D$ matrices
($D$ being the size of the readout layer), and they modify the NNGP kernel via a Kronecker product.
This process is called kernel shape renormalization (\citealp{SompolinskyLinear, pacelli2023statistical}).
Yet another transformation of the kernel, coined local kernel renormalization, was found for
convolutional networks  in \cite{aiudi2023}, where the NNGP kernel splits into several local components (\citealp{novak2019bayesian}), labelled by two additional indices $i,j$ that run across the patches in each element of the training set (e.g., the receptive fields of an image).
In the proportional limit, each local component $(i,j)$ of the kernel is modified by an order parameter $Q_{ij}$. 
An interesting type of kernel shape renormalization was found in
globally gated deep linear networks too (\citealp{li2022globally}), but the interpretation is different than in the convolutional case presented in \cite{aiudi2023}.

\begin{table}[h]
    \centering
    \begin{tabular}{c|l}
    \multicolumn{2}{c}{\textbf{Common definitions}}\\
      $P$   &  size of the training set\\
      $L$   &  depth of the network\\
      $N_0$ &  input dimension\\
      $\theta$ & collection of all trainable weights, $\theta = \{W^{(\ell)} \}_{\ell=0}^L$\\
      $\mathbf{x}^\mu$, $X$ & input of the network in the training set, $X = [\mathbf{x}^\mu]_{\mu=1}^P$ \\
      $\by^\mu$, $\by_{1:P}$ & label (response) of $\mathbf{x}^\mu$ in the training set, $\by_{1:P}^\top=(\by^{1 \top},\dots,\by^{P\top})$\\
      $f_\theta(\mathbf{x}^\mu)$ & output of the network from input $\mathbf{x}^\mu$\\
      \multicolumn{2}{c}{\rule{0pt}{4ex}\textbf{Fully-connected}}\\
      $N_1,\cdots,N_{L}$ & width of each hidden layer\\
      $N_{L+1}=D$ &  output dimension \\
      $\mathbf{S}^\mu$, $\mathbf{S}_{1:P}$ & output of the network in $\mathbb{R}^D$, $\mathbf{S}^\mu = f_\theta(\mathbf{x}^\mu)$, $\mathbf{S}_{1:P}^\top=(\mathbf{S}^{1 \top},\dots,\mathbf{S}^{P\top})$\\
      \multicolumn{2}{c}{\rule{0pt}{4ex}\textbf{Convolutional}}\\
      $C_0,\cdots,C_L$ & number of channels in each hidden layer\\
      $M$ & dimension of the channel mask\\
       $S^\mu$, $\mathbf{S}_{1:P}$ & output of the network in $\mathbb{R}$, $S^\mu = f_\theta(\mathbf{x}^\mu)$, $\mathbf{S}_{1:P}^\top=(S^{1},\dots,S^{P})$
    \end{tabular}
    \caption{Table of notations}
    \label{tab:notations}
\end{table}

\section{Problem setting}

We consider a supervised learning problem with training set $\{\mathbf x^\mu ,\mathbf y^\mu\}_{\mu=1}^P$, where each $\mathbf x^\mu \in \mathbb R^{N_0 \times C_0}$ and the corresponding labels (response) $\mathbf y^\mu \in \mathbb{R}^{D}$, with $C_0=1$ for
the fully-connected architecture and, for simplicity, $D=1$ for the convolutional architecture. 
The output will be denoted by $f_{\theta} (\mathbf x^\mu)$, where $\theta$ represents the collection of all the trainable weights of the network (see Table~\ref{tab:notations} for a summary of notations). 

\subsection{Fully-connected deep linear neural networks (FC-DLNs)}

In fully-connected neural networks, 
the pre-activations of each layer $h_{i_{\ell}}^{(\ell)}$ ($i_{\ell} = 1,\dots, N_{\ell}$; $\ell = 1, \dots, L$) are given recursively as a function of the pre-activations of the previous layer $h_{i_{\ell-1}}^{(\ell-1)}$ ($i_{\ell-1}= 1, \dots, N_{\ell-1}$):
\begin{equation}
    \label{main_recursion}
    \begin{split}
h_{i_1}^{(1)}(\mathbf x^\mu) &= \frac{1}{\sqrt {N_{0}}} \sum_{i_0=1}^{N_{0}} W^{(0)}_{i_1 i_0} x_{i_0}^\mu + b_{i_1}^{(0)}\,, \\
h_{i_\ell}^{(\ell)}(\mathbf x^\mu) &= \frac{1}{\sqrt {N_{\ell-1}}} \sum_{i_{\ell-1}=1}^{N_{\ell-1}} W^{(\ell-1)}_{i_{\ell}i_{\ell-1}} h_{i_{\ell-1}}^{(\ell-1)}(\mathbf x^\mu) + b_{i_\ell}^{(\ell-1)}\,, \\
\end{split}
\end{equation}
where $W^{(\ell-1)}$ and $b^{(\ell-1)}$ are respectively the weights and the biases of the $\ell$-th layer, whereas the input layer has dimension $N_0$ (the input data dimension). 
In what follows we shall take $b_{i_\ell}^{(\ell-1)}=0$ for every $\ell$
and every $i_\ell$ (without loss of generality, as one can include this contribution in the vectors of weights adding additional dimensions in input and hidden space).

We add one last readout layer with 
dimension $N_{L+1}=D$, and define the function implemented by the deep neural network as
$f_{\theta} (\mathbf x^\mu)=(S^\mu_1,\dots,S^\mu_D)^\top$, where
\begin{equation}
\label{eq:f_DNN}
S^{\mu}_i  = 
\frac{1}{\sqrt {N_{L}}} \sum_{i_{L}=1}^{N_{L}} W^{(L)}_{i,i_{L}} h_{i_{L}}^{(L)}(\mathbf x^\mu)
\qquad i=1,\dots,D\, ; \,\, \mu=1,\dots,P,
\end{equation}
 $W^{(L)}_{i,i_{L}}$ being the  weights of the last layer. 
The input training set is collected 
in a  $N_0 \times P$  matrix $\bX$, that is 
\[
\bX=[\bx^{1},\dots,\bx^{P}] , 
\]
and the corresponding labels in a vector 
$\by_{1:P}^\top=(\by_1^\top,\dots,\by_P^\top)$.
We also denote by
\[
\bS_{1:P}=(S^{1}_1,\dots,S^{1}_D,S^2_1,\dots,S^P_1,\dots,S^P_D)^\top
\]
the $P$ outputs
$(\bS^{\mu}=f_{\theta} (\mathbf x^\mu): \mu=1,\dots,P)$ stacked in a vector. 

\subsection{Convolutional deep linear neural networks (C-DLNs)}

For convolutional neural networks,  
 the pre-activations at each layer $\ell = 1, \dots, L$ are labelled by two indices: (i) the channel index $a_\ell = 1, \dots,C_\ell$, where $C_\ell$ is the total number of channels in each layer; (ii) the spatial index $i =1, \dots, N_0$, which runs over the input coordinates. 
 Similarly any input has two indices, that is
 $\mathbf x^\mu=[x^\mu_{a_{0},i}]$, with $a_0=1,\dots,C_0$ and
 $i-1,\dots,N_0$. 
Pre-activations of the first layer are given by 
\begin{equation}
    h_{a_1,i}^{(1)}(\mathbf x^\mu) = \frac{1}{\sqrt{M C_{0}}} \sum_{a_{0}=1}^{C_{0}} \sum_{m = -\lfloor M/2 \rfloor}^{\lfloor M/2 \rfloor} W^{(0)}_{m,a_1 a_{0}} x^\mu_{a_{0},i+m  (\mathrm{mod} N_0)}
\end{equation}
and for $\ell>1$ 
\begin{equation}\label{convolutiona.recur1}
    h_{a_\ell,i}^{(\ell)}(\mathbf x^\mu) = \frac{1}{\sqrt{M C_{\ell - 1}}} \sum_{a_{\ell - 1}=1}^{C_{\ell - 1}} \sum_{m = -\lfloor M/2 \rfloor}^{\lfloor M/2 \rfloor} W^{(\ell-1)}_{m,a_\ell a_{\ell -1}} h^{(\ell - 1)}_{a_{\ell-1},i+m (\mathrm{mod} N_0)}(\mathbf x^\mu)\,,
\end{equation}
where $M$ is the dimension of the channel mask. 
Here we  restrict our analysis to $1d$ convolutions, periodic boundary conditions over the spatial index and unitary stride. This choice allows to keep the size of the spatial index constant through layers. Note that in the case of non-unitary stride, one should introduce a different spatial index $i_\ell$ for each layer, which runs from 1 to the integer part of 
the ratio between $N_0$ and the stride's $\ell$th power. 
The output of the network is given by
\begin{equation}
    S^\mu = f_\theta(\mathbf{x}^\mu) = \frac{1}{\sqrt{C_L N_0}} \sum_{a_L=1}^{C_L} \sum_{i=1}^{N_0} W^{(L)}_{a_L i} h^{(L)}_{a_L,i}(\mathbf{x}^\mu)\,.
    \label{eq:CNN_S}
\end{equation}
Again we denote by $\bS_{1:P}$ the vector of outputs $(S^1,\dots,S^P)$
and the data is now collected in a three-dimensional array
$\bX=[x^\mu_{a_{0},i}]_{\mu,a_0,i}$. 

\subsection{Bayesian setting}\label{Sec:BayesianStting}

In a Bayesian neural network, 
 a prior for the weights $\theta$ is specified, which 
translates in a prior for $f_{\theta} (\mathbf x^\mu)$. 
In other words, the prior (over the outputs) is the density of the random vector $\bS_{1:P}$.
This density 
evaluated for a given value $\bs_{1:P}$ taken by $\bS_{1:P}$, will be denoted by $p_{\mathrm{prior}}(\bs_{1:P}|\bX )$. 
At this stage,  one needs also a likelihood  for the 
labels given the inputs and the outputs, 
denoted by $\CL(\by_{1:P}|\bs_{1:P},\beta)$, where $\beta$ is an additional parameter, which plays the role of the inverse temperature in statistical physics. 
The posterior density of $\bS_{1:P}$ 
given $\by_{1:P}$ is (by Bayes theorem) 
 \[
p_{\mathrm{post}}(\bs_{1:P}|\by_{1:P},\bX)  \propto 
p_{\mathrm{prior}}(\bs_{1:P}|\bX) \CL( \by_{1:P}|\bs_{1:P}).
\]
In order to describe the 
posterior predictive, 
let $\mathbf x^0$ be a new input and  set $\bS^0= f_{\theta} (\mathbf x^0)$.
The  posterior predictive density of $\bS_0$  given $\by_{1:P}$
   is obtained similarly  by
 \[
p_{\mathrm{pred}}(\bs_0|\by_{1:P},\bx^0,\bX)  := Z_{\by_{1:P},\beta}^{-1}
\int_{\RE^{D\times P}} \CL( \by_{1:P}|\bs_{1:P})
p_{\mathrm{prior}}(\bs_0,\bs_{1:P}|\bx^0,\bX) d\bs_{1:P}
\]
where 
$p_{\mathrm{prior}}(\bs_0,\bs_{1:P}|\bx^0,\bX)$ is the prior 
for $(\bS_0,\bS_{1:P})$ and 
\begin{equation}\label{normalization_Z}
Z_{\by_{1:P},\beta}
=\int_{\RE^{D(P+1)}} \CL( \by_{1:P}|\bs_{1:P})
p_{\mathrm{prior}}(\bs_0,\bs_{1:P}|\bx^0,\bX)d\bs_{0} d\bs_{1:P}
\end{equation}
is the partition function (normalization constant). In the 
convolutional setting we have $D=1$ and, consequently, we shall write $s_0$ in place of $\bs_0$. 

Note that, training the network with  a quadratic loss function is equivalent to consider a Gaussian likelihood in the Bayesian setting, that is 
\[
\CL(\by_{1:P}|\bs_{1:P},\beta)  \propto e^{-\frac{\beta}{2} \|\bs_{1:P}-\by_{1:P}\|^2}.
\]
 Note that in this case 
\[
\by^\mu=\bS^\mu+\bm{\epsilon}^\mu=f_\theta(\bx^\mu)+\bm{\epsilon}^\mu \quad \bm{\epsilon}^\mu \stackrel{iid}{\sim} \CN(0,\beta^{-1}\eI_D).
\]
Here, we 
assume that the weights are independent  normally distributed with zero means 
and variance $1/\lambda_{\ell}$ at layer $\ell$. In summary:
\begin{itemize}
\item 
For the fully connected network (FC-DLN):  
\begin{equation}\label{lawofW}
   W^{(\ell)}_{ij}  \stackrel{iid}{\sim} \CN(0,\lambda_\ell^{-1})   
 \end{equation}
\item  For the convolutional  network (C-DLN): 
\begin{equation}
    W^{(\ell)}_{m,a b} \overset{iid}{\sim} \mathcal{N}(0,\lambda_\ell^{-1})\,,\qquad W^{(L)}_{a i} \overset{iid}{\sim} \mathcal{N}(0,\lambda_L^{-1})\,.
\end{equation}
\end{itemize}

\section{Results}

Our main results are collected in this section. 

\subsection{Exact non-asymptotic integral representation for the prior over the outputs}

We start discussing the FC-DLN. To state the results, denote by $\CS^+_{D}$ the set of  symmetric
 positive definite matrices 
 on $\RE^{D}$ and recall 
 that a random matrix  $Q$ taking values in  $\CS^+_D$ has 
 Wishart  distribution with $N>D$ 
 degrees of freedom and scale matrix $V$  if it has the following density
(with respect to  the Lebesgue measure on the cone of symmetric positive definite matrices)
\[
 \CW_D(Q|V,N)=\frac{\det(Q)^{\frac{N-D-1}{2} }  e^{-\frac{1}{2}\tr(V^{-1}Q)}}{\det(V)^{N/2}2^{DN/2}(\pi)^{D(D-1)/4}\prod_{k=1}^D\Gamma\!\left(\frac{N - D +k}{2}\right)}.
 \]
Equivalently, $Q$ has 
a Wishart distribution with $N$ degrees of freedom and scale matrix $V$ if its law corresponds to the law of 
$\sum_{i=1}^N \bZ_i \bZ_i^\top$ where $\bZ_i$ are independent   Gaussian vectors in $\RE^D$
with zero mean and  covariance matrix $V$ (see~\citealp{Eaton2007} and the references therein).  
If $D=1$, one has that $\CS^+_{D}=\RE^+$ and the Wishart distribution reduces to  a gamma distribution of parameters $\alpha=\frac{N}{2}$, $\beta=\frac{1}{2V}$, 
that is 
\[
\CW_1(Q|V,N)=
\frac{1}{(2V)^{N/2} \Gamma(N/2)}   
Q^{\frac{N}{2}-1} e^{- \frac{1}{2V}Q} \quad \text{for $Q>0$}.
\]
In our integral representations a fundamental role is played by 
 the joint distribution 
 of the vector of matrices 
 \[
 (Q_1,\dots,Q_L):=(\tilde Q_1/ N_1,\dots, \tilde Q_L/N_L),
 \]
 where 
 the $\tilde Q_\ell$'s are $D \times D$ independent Wishart random matrices with $N_\ell$ degree of freedoms and identity scale matrix
 $\eI_D$. 
This joint distribution of $(Q_1,\dots,Q_L)$ is
\begin{equation}\label{def:QLN}  
\mathcal{Q}_{L,N} (d  Q_1 \dots d  Q_L)
:=\left ( \prod_{\ell=1}^L \CW_D\Big(Q_\ell\Big|\frac{1}{N_\ell }\eI_D,N_\ell\Big)  \right) 
dQ_1 \dots dQ_L
\end{equation}
 For the sake of notation, we  write  
$\mathcal{Q}_{L,N}$ in place of the more
correct $\mathcal{Q}_{L,N_1,\dots,N_L}$.

We now show that the prior distribution over $\bS_{1:P}$ is a  mixture of Gaussians where the covariance matrix is
an explicit function of random Wishart matrices.
In what follows $A \otimes B$ denotes the 
Kronecker product of matrices $A$ and $B$. Moreover,  
we set 
$\lambda^*:=\lambda_0 \dots \lambda_L$. 

\begin{proposition}\label{prop1}
{\bf(non-asymptotic integral representation for the prior of FC-DLNs with multiple outputs)}
Let $\hat p_{\mathrm{prior}}$ be the characteristic function
of the outputs $\bS_{1:P}$ of a 
FC-DLN, that is   
$\hat p_{\mathrm{prior}}(\bar \bs_{1:P}|\bX )
=\E[\exp\{i \bar \bs_{1:P}^\top \bS_{1:P} \}]$. 
If $\min(N_\ell:\ell=1,\dots,L) >D$, then 
\[
\hat p_{\mathrm{prior}}(\bar \bs_{1:P}|\bX )
= \int_{(\CS^+_{D})^L}  \!\!\!
e^{-\frac{1}{2} \bar\bs_{1:P}^\top \FCK(Q_1,\dots,Q_L) \bar\bs_{1:P} }
\mathcal{Q}_{L,N}(d Q_1\dots dQ_L)
\]
where 
$\mathcal{Q}_{L,N}$ is 
defined by \eqref{def:QLN}, 
$\FCK(Q_1,\dots,Q_L)=(N_0\lambda^*)^{-1}\bX^\top \bX \otimes Q^{(L)}(Q_1,\dots,Q_L)$,
and 
\[
Q^{(L)}=Q^{(L)}(Q_1,\dots,Q_L):=
( U_1 \dots  U_{L})^\top U_1 \dots  U_{L},
\]
 $U_\ell$ being $D \times D$ matrices  such that 
 $U_{\ell}^\top U_{\ell}=Q_\ell$.  
\end{proposition}

\begin{remark}
In the previous representation 
one can choose any decomposition 
of $Q_\ell$ in $U_\ell^\top  U_\ell$. For convenience   in what follows  we choose the Cholesky decomposition 
where $U_\ell^\top= U_\ell^\top( Q_\ell)$ is a lower triangular matrix with positive diagonal entries. This $U_\ell^\top$ is called Cholesky factor (or Cholesky square root).
 The Cholesky square root 
 is one to one and continuous
from $\CS_D^+$ to its image. 
When $Q_\ell$ has a Wishart distribution the law of it
Cholesky square root $U_\ell^\top$ is the so-called
Bartlett distribution (see~\citealp{kshirsagar1959}).
\end{remark}

\begin{remark}\label{Rem2}
The integral representation stated above is equivalent to say 
that  the random matrix 
$ S=[S^{\mu}_d]_{d=1,\dots,D;\mu=1,\dots,P}$ 
admits the explicit  stochastic representation 
\begin{equation}\label{main_in_matrix_form}   
  S = U_{L}^\top \cdots  U_{1}^\top \frac{Z }{\sqrt{N_0 \lambda^*}}   \bX
\end{equation}
where $U_{\ell}$ are $D \times D$ independent random matrices
such that $Q_\ell= U_{\ell}^\top U_{\ell}$
  has a Wishart distribution 
with $N_\ell$ degrees of freedom 
and scale matrix  $\frac{1}{N_\ell}\eI_D$ and 
 $Z$ is a $D \times N_0$ matrix of independent standard normal random variables. 
 As mentioned in the introduction, all the $N_\ell$'s appear parametrically in the above representation, 
 providing the anticipated {dimensional reduction}. 
 For comparison, note that  
 by \eqref{main_recursion} one can derive  another 
 explicit explicit expression for $S$, namely
\begin{equation}\label{basic_rep_in_matrix_form}
    S =  \frac{W^{(L)}}{\sqrt{N_{L}}} \cdots
    \frac{W^{(0)}}{\sqrt{N_{0}}}
    \bX   
\end{equation}
where $W^{(\ell)}=[W_{i_{\ell+1},i_\ell}^{(\ell)} ]$ are $N_{\ell+1}\times N_{\ell}$ matrices 
of normal random variables of zero mean and variance $1/\lambda_\ell$.
This second representation is more direct (and easier to derive)
and also shows that $S$ is a  mixture of Gaussians.
Nevertheless, containing  the $N_{\ell+1}\times N_{\ell}$ matrices $W^{(\ell)}$, it 
does not provide any dimensional reduction. 
\end{remark}

\begin{remark}
As recalled above, for $D=1$ the $U_{\ell}$'s 
are independent random variables
 distributed as the square root of a $\GGamma(N_{\ell}/2, N_{\ell}/2)$ 
 random variable, that is $\tilde Q_\ell=N_{\ell} {U_{\ell}}^2$
is a $\chi^2$ with ${N_{\ell}}$ degrees of freedom. 
The law of a product of independent gamma random variables
is derived in~\citet{springer1970products}. 
Using this result, when $D=1$,
one gets that 
the rescaled random variable 
 $\tilde{Q}^{(L)} = \prod_{i=1}^L (2\tilde{Q}_\ell /N_\ell)$ 
 has density 
\begin{equation}\label{dens_Meijer}
    p_{\tilde{Q}^{(L)}}(q) = \frac{1}{\prod_{\ell = 1}^L \Gamma(N_\ell/2) } G^{L0}_{0L}\!\left(q\,\, \Big | \begin{matrix} - \\ \frac{N_1 }{2} - 1, \cdots, \frac{N_L}{2} - 1  \end{matrix}\right)\,,
\end{equation}
where $G^{m,n}_{p,q}$ is the Meijer G-function. 
In the case of Gaussian likelihood, one can employ \eqref{dens_Meijer} to give an alternative derivation of the non-asymptotic expression of the partition function (Bayesian model evidence) given by \cite{doi:10.1073/pnas.2301345120}.
We provide more  details on this derivation in Section 
\ref{Meijer-Appendix}.
 Note that the link between Meijer G-functions and the deep linear networks based on gamma functions was first established in \cite{zavatone-veth2021exact} in the limited setting of individual training patterns (i.e. for the prior over a single input vector). 
\end{remark}

While in  \cite{doi:10.1073/pnas.2301345120} the partition function is given in terms of Meijer G-functions, here it is expressed as an $L-$dimensional integral over a set of order parameters $Q_\ell$. 
Using our alternative representation, it is possible to compare and find some equivalence with the results (in the proportional limit $P, N_\ell \to \infty$ at fixed $\alpha_\ell = P/N_\ell$) found in  \cite{SompolinskyLinear}
and \cite{pacelli2023statistical} (see the Discussion in Section~\ref{sec3}).

In order to discuss a non-asymptotic representation formula for the convolutional 
architecture, it is convenient to introduce the translation operator $T_m$ as
\begin{equation}
    T_{m,ij} = \delta_{j,i+m (\mathrm{mod} N_0)}.
\end{equation} 
In this way, Eq. \eqref{convolutiona.recur1} takes the following form  
\begin{equation}\label{convolutiona.recur2}
    h_{a_\ell,i}^{(\ell)}(\mathbf x^\mu) = \frac{1}{\sqrt{M C_{\ell - 1}}} \sum_{a_{\ell - 1}=1}^{C_{\ell - 1}} \sum_{m = -\lfloor M/2 \rfloor}^{\lfloor M/2 \rfloor} W^{(\ell-1)}_{m,a_\ell a_{\ell -1}} \sum_{j=1}^{N_0}T_{m,ij} h^{(\ell - 1)}_{a_{\ell-1},j}(\mathbf x^\mu)\,.
\end{equation}
One can now define a map $\mathcal{T}: (\CS^+_{N_0})^L \to \CS^+_{N_0}$
that will play a role similar to the map 
 $Q^{(L)}=Q^{(L)}(Q_1,\dots,Q_L)$ 
 introduced  in {\rm Proposition~\ref{prop1}}.
Given $Q_1,\dots,Q_L$, one sets $\mathcal{T}(Q_1,\dots,Q_L)=
 Q_1^*$ where $Q_L^*,\dots, Q_1^*$ are defined  by the following backward recursion:
\begin{equation}
\label{eq:CNN_Qstar}
\begin{aligned}
Q^*_L &= \frac{1}{M} \sum_{m = -\lfloor M/2 \rfloor}^{\lfloor M/2 \rfloor} T^\top_m Q_{L} T_m\,, \\
    {Q}_{\ell}^* &=\frac{1}{M}\sum_{m = -\lfloor M/2 \rfloor}^{\lfloor M/2 \rfloor} T^\top_m
    \left( (U_{\ell+1}^*)^\top Q_{\ell}  U_{\ell+1}^* \right) T_m
    \quad \text{for $\ell=L-1,\dots,1$}
\end{aligned}
\end{equation}
 $U_{\ell+1}^*$ being a square root of $Q^*_{\ell+1}$,
that is $(U_{\ell+1}^*)^\top U_{\ell+1}^*= Q_{\ell+1}^*$. 
Note that   $[T_{m}^\top QT_{m}]_{i,j}=Q_{i+m (\text{mod} N_0),j+m (\text{mod} N_0)}$
for any matrix $Q$.
Finally, we need the analogous of $\mathcal{Q}_{L,N}$ for the convolutional architecture. To this end,  we set
\begin{equation}\label{def:QLN_CNN}  
\mathcal{Q}_{L,C} (d  Q_1 \dots d  Q_L)
:=\left ( \prod_{\ell=1}^L \CW_{N_0}\Big(Q_\ell\Big|\frac{1}{C_\ell }\eI_{N_0},C_\ell\Big)  \right) 
dQ_1 \dots dQ_L.
\end{equation}
 Note that $\mathcal{Q}_{L,C}$ has exactly the same 
form of $\mathcal{Q}_{L,N}$ once $N_1,\dots,N_L$
is replaced by $C_1,\dots,C_L$. 

\begin{proposition}\label{prop1-CNN}
{\bf (non-asymptotic integral representation for the prior of C-DLNs with single output)}
For a  C-DLN with $\min(C_\ell:\ell=1,\dots,L) >N_0$, 
the characteristic function 
$\hat p_{\mathrm{prior}}(\bar \bs_{1:P}|\bX )
=\E[\exp\{i \bar \bs_{1:P}^\top \bS_{1:P} \}]$
of the outputs $\bS_{1:P}$ 
is given by 
\[
\hat p_{\mathrm{prior}}(\bar \bs_{1:P}|\bX ) 
= \int_{(\CS^+_{N_0})^L} 
e^{-\frac{1}{2} \bar \bs_{1:P}^\top (\CNN(Q_1,\dots,Q_L)) \bar\bs_{1:P} }\mathcal{Q}_{L,C}(d Q_1\dots dQ_L)
\]
where 
$\mathcal{Q}_{L,N}$ is 
defined by \eqref{def:QLN_CNN}, 
and the covariance matrix $\CNN(Q_1,\dots,Q_L)$
is defined as 
\begin{equation}\label{Def_CNN_matrix}
    \CNN(Q_1,\dots,Q_L):=
   \Big [ \sum_{r,s}\mathcal{T}(Q_1,\dots,Q_L)_{rs} \sum_{a_0} \frac{x^{\mu}_{a_0,r} x^{\nu}_{a_0,s}}{\lambda^* C_0} \Big]_{\mu=1,\dots,P; \nu=1,\dots, P}.
\end{equation}
\end{proposition}

Note that the map $\mathcal T$ acts as on a four indices tensor, providing a partial trace wrt to indices $r,\, s$. This tensor is known in the literature as \emph{local covariance matrix} from the seminal work \cite{novak2019bayesian} on the Bayesian infinite-width limit in convolutional networks. The same partial trace structure has been also identified in the proportional limit for non-linear one-hidden-layer networks in \cite{aiudi2023}, where the authors coined the term \emph{local kernel renormalization} to indicate the re-weighting of the infinite-width local NNGP kernel found in \cite{novak2019bayesian}.

\subsection{Posterior distribution and predictor's statistics in the case of squared error loss function}

In the case of quadratic loss function (Gaussian likelihood), it is possible to show that the predictive posterior of $\bS_0=f_{\theta} (\mathbf x^0)$ is a also  
mixture of Gaussian distributions, where the mixture is now
both on the covariance matrix and on the vector of means and 
it is driven by an updated version of $\mathcal{Q}_{L,N}$.
In the next result we assume that the matrix
$\tilde\bX=[\bx^{0},\bx^{1},\dots,\bx^{P}]$
is full rank.

\begin{proposition}\label{Prop2}
Let 
$\mathcal{Q}_{L,N}(dQ_1 \dots dQ_L)$ 
be as in Eq. \eqref{def:QLN} 
and $Q^{(L)}=Q^{(L)}(Q_1,\dots,Q_L)$ 
be defined as in {\rm Proposition~\ref{prop1}}.
Let  $p_{\mathrm{pred}}(\bs_0|\by_{1:P},\bx_0,\bX)$
be the posterior predictive of  
the FC-DLN and assume that 
 $\det(\tilde \bX^\top \tilde\bX)>0$. 
Then  
for every $\bs_0 \in \RE^{D}$ one has
\[
\begin{split}
& p_{\mathrm{pred}}(\bs_0|\by_{1:P},\bx_0, \bX)  \\
& \quad 
= \int_{(\CS^+_{D})^L}  \!\! \frac{
e^{- \frac{1}{2}(\bs_0- \mathbf{m}_0)^\top  (\Sigma_{00}- 
\Sigma_{01}(\Sigma_{11}+\beta^{-1}\eI_{D P})^{-1} \Sigma_{01}^\top)^{-1} (\bs_0-\mathbf{m}_0) }
}{(2 \pi)^{\frac{D}{2}}\det(  
\Sigma_{00}- 
\Sigma_{01} (\Sigma_{11}+\beta^{-1}\eI_{D P})^{-1} \Sigma_{01}^\top)^{\frac{1}{2}}  }
 \mathcal{Q}_{L,N}(dQ_1 \dots dQ_L| \by_{1:P},\beta) \\
 \end{split}
\]
where:
\[
\begin{split}
& \Sigma_{00}=  \frac{\bx_0^\top \bx_0}{N_0 \lambda^*}\otimes Q^{(L)}, \quad
\Sigma_{01}= \frac{\bx_0^\top \bX}{N_0 \lambda^*}\otimes Q^{(L)}, \quad  
\Sigma_{11}=\frac{\bX^\top \bX}{N_0 \lambda^*} \otimes Q^{(L)} 
\\
& 
\mathbf{m}_0= \Sigma_{01} ( \Sigma_{11} +\beta^{-1} \eI_{D P})^{-1}
\by_{1:P},
\\
\end{split}
\]
\[  
\mathcal{Q}_{L,N}(dQ_1 \dots dQ_L| \by_{1:P},\beta)   :=
 \frac{e^{-\frac{1}{2} \Phi_\beta(Q_1 \dots Q_L,\by_{1:P}) }\mathcal{Q}_{L,N}(dQ_1 \dots dQ_L)  }{\int_{(\CS^+_{D})^L}e^{-\frac{1}{2} \Phi_\beta(Q_1 \dots Q_L,\by_{1:P}) } \mathcal{Q}_{L,N}(dQ_1 \dots dQ_L) }   
 \]
 and  $\Phi_\beta(Q_1 \dots Q_L,\by_{1:P}):=
\by_{1:P}^\top ( \Sigma_{11} +\beta^{-1}  \eI_{D P})^{-1} \by_{1:P}
 +\log(\det( \eI_{D P} +\beta \Sigma_{11}))$.
\end{proposition}

An analogous result holds for C-DLN, 
where now the output is in dimension $1$. 
Here we need to define 
$\tilde \bX=[x^\mu_{a_{0},i}]_{\mu,a_0,i}$
where now  $\mu=0,\dots,P$
and, for every $a_0=1,\dots,N_0$,  the matrix $\tilde \bX_{a_0}=[x^\mu_{a_{0},i}]_{\mu,i}$.

\begin{proposition}\label{Prop2-CNN}
Let 
$\mathcal{Q}_{L,C}(dQ_1 \dots dQ_L)$ 
be defined in Eq. \eqref{def:QLN_CNN}
and  $p_{\mathrm{pred}}(s_0|\by_{1:P},\bx_0,\bX)$
be the posterior predictive of  
the C-DLN.  Assume that 
$\det(\sum_{a_0=1}^{N_0} \tilde \bX_{a_0}^\top \tilde \bX_{a_0})>0$.
Then,  for any $s_0 \in \RE$ one has 
\[
\begin{split}
& p_{\mathrm{pred}}(s_0|\by_{1:P},\bx_0, \bX)  \\
& \quad = \int_{(\CS^+_{N_0})^L} \frac{
e^{- \frac{1}{2}  (\Sigma_{00}- 
\Sigma_{01} (\Sigma_{11}+\beta^{-1}\eI_{P})^{-1} \Sigma_{01}^\top)^{-1} (s_0- m_0)^2}}{(2 \pi)^{\frac{1}{2}}
\det(  
\Sigma_{00}- 
\Sigma_{01} (\Sigma_{11}+\beta^{-1}\eI_{P})^{-1} \Sigma_{01}^\top)^{\frac{1}{2}}  }
 \mathcal{Q}_{L,C}(dQ_1 \dots dQ_L| \by_{1:P},\beta)  \\
 \end{split}
\]
where:
\[
\begin{split}
& \Sigma_{00}:= \sum_{r,s}\mathcal{T}(Q_1,\dots,Q_1)_{rs} \sum_{a_0} \frac{x^{0}_{a_0,r} x^{0}_{a_0,s}}{\lambda^* C_0}, \\
& \Sigma_{01}:=\Big[\sum_{r,s}\mathcal{T}(Q_1,\dots,Q_1)_{rs} \sum_{a_0} \frac{x^{\mu}_{a_0,r} x^{0}_{a_0,s}}{\lambda^* C_0}\Big]_{\mu=1,\dots,P}
\\
& \Sigma_{11}:= \Big[\sum_{r,s}\mathcal{T}(Q_1,\dots,Q_1)_{rs} \sum_{a_0} \frac{x^{\mu}_{a_0,r} x^{\nu}_{a_0,s}}{\lambda^* C_0}
   \Big]_{ \mu,\nu=1,\dots,P}, \\
& m_0:= \Sigma_{01} ( \Sigma_{11} +\beta^{-1} \eI_{ P})^{-1}
\by_{1:P},
\\
& \mathcal{Q}_{L,C}(dQ_1 \dots dQ_L| \by_{1:P},\beta)   :=
 \frac{e^{-\frac{1}{2} \Phi_\beta(Q_1 \dots Q_L,\by_{1:P}) }\mathcal{Q}_{L,N}(dQ_1 \dots dQ_L)  }{\int_{(\CS^+_{N_0})^L}e^{-\frac{1}{2} \Phi_\beta(Q_1 \dots Q_L,\by_{1:P}) } \mathcal{Q}_{L,C}(dQ_1 \dots dQ_L) }  
 \\
\end{split}
\]
 and  $\Phi_\beta(Q_1 \dots Q_L,\by_{1:P}):=
\by_{1:P}^\top ( \Sigma_{11} +\beta^{-1}  \eI_{P})^{-1} \by_{1:P}
 +\log(\det( \eI_{P} +\beta \Sigma_{11}))$.
\end{proposition}

\subsection{Quantitative description of the feature learning infinite-width regime}

In this section, we give a quantitative description of the feature learning infinite-width regime, using large deviation theory. For concreteness, we specialize to the case of FC-DLNs with multiple outputs, but analogous statements hold for C-DLNs.

Following the literature (\citealp{ChizatLazy, doi:10.1073/pnas.1806579115, GEIGER20211, Geiger_2020, yang2020feature}), we recall that the easiest way to escape the lazy-training infinite-width limit is to consider the so-called \emph{mean-field parametrization}, which corresponds to the following rescaling of the loss and output functions:
\begin{equation}
    \CL_N(\by_{1:P}| \bs_{1:P},\beta) :=\CL(\by_{1:P}| \bs_{1:P}/\sqrt{N},N \beta).
    \label{GLikeFL}
\end{equation}
 We point out that the mean-field parameterization exhibits some pathological behavior in the Bayesian setting.
In a sense, the scale of the prior is incorrect, since it looks like a delta function in the limit. Nevertheless, one can compute the posterior of the random covariance appearing in the mixture representation, and this posterior exhibits a well-defined and non-trivial limiting behavior. 
Comparing the large deviation asymptotics of the mean-field  posterior covariance with those in the lazy-training infinite-width limit, one recognizes the presence of additional terms, which can be interpreted as an instance of feature learning.

As starting point, we briefly consider the lazy-training infinite-width limit. 
Here $L$ and $D$ are fixed and 
$N_1=\dots,N_L=N$ and 
$N \to +\infty$.
 It is easy to see that in this case
$(\tilde Q_1/(\lambda_1 N_1),\dots, \tilde Q_L/(\lambda_L N_L))$, converges (almost surely) 
to $(\eI_D,\dots,\eI_D)$.
This is a consequence of the law of large numbers, indeed 
 $Q_\ell=\frac{\tilde Q_\ell}{N}=\frac{1}{N} \sum_{j=1}^N 
\bZ_{j,\ell} \bZ_{j,\ell}^\top$ with $\bZ_{j,\ell}$
independent standard Gaussian's vectors. 
Since the map $(Q_1,\dots,Q_L) \mapsto Q^{(L)}$ in Proposition~\ref{prop1} is  continuous, also $Q^{(L)}$ converges almost surely to $\eI_D$. In this way, one recovers the well-known Gaussian lazy-training infinite-width limit.

For $L$  fixed, it is easy to  derive  a 
large deviation principle (LDP) for the random 
covariance matrix $Q^{(L)}$ appearing 
in the prior representation given in Proposition 
\ref{prop1} or, equivalently, 
for $\mathcal{Q}_{L,N}$.

Roughly speaking, if $V_N \in \RE^D$ satisfies an LDP with rate function $I$ then 
$P\{V_N \in dx\}=P_N(dx) \simeq e^{-N I(x)} dx$. 
The formal definition 
(see, e.g.,  \citealp{DemboZeitouni}) is as follows:
a sequence of random elements 
$(V_N)_N$ taking values in a metric space $(\mathbb{V},d_V)$ (or equivalently 
the sequence of their laws, say $P_N$) satisfies 
an LDP with rate function $I$ if 
\[
\limsup_{N \to +\infty}
\frac{1}{N} \log \Big(
\mathbb{P}\{ V_N \in C\}\Big )
=\limsup_{N \to +\infty} \frac{1}{N} \log\Big (
P_N(C) \Big)
\leq -\inf_{v \in C} I(v) 
\]
for any closed set $C \subset \mathbb{V}$
 and  
\[
\liminf_{N \to +\infty}\frac{1}{N}
\log \Big ( \mathbb{P}\{ V_N \in O\}\Big )
=\liminf_{N \to +\infty}\frac{1}{N}
\log \Big ( P_N(O)\Big )
\geq -\inf_{v \in O} I(v)
\]
for any open set $O \subset \mathbb{V}$.

\begin{proposition}\label{LDPeasy}
The sequence of measures $\mathcal{Q}_{L,N}$,
satisfies an LDP  
on $({\CS^+_D})^L$ as $N \to +\infty$ ($L$ fixed) 
with rate  function
\[
I(Q_1,\dots,Q_L):=\frac{1}{2}
\sum_{\ell=1}^L \Big(  \tr(Q_\ell)-
\log(\det(Q_\ell)\Big)-\frac{DL}{2}.
\]
Moreover, the sequence of measures   $\mathcal{Q}_{L,N}(dQ_1 \dots dQ_L | \by_{1:P},\beta) $ 
 defined  in {\rm Proposition~\ref{Prop2}}, 
 satisfies an LDP with the same rate function, 
 independently from $\beta$ and $\by_{1:P}$. 
\end{proposition}

\begin{remark}
As a corollary, using the well-known 
contraction principle (see, e.g., Theorem 4.2.1 in \citealp{DemboZeitouni}), 
if $(Q_1,\dots,Q_L)$
has law $\mathcal{Q}_{L,N}$, then 
$Q^{(L)}(Q_1,\dots,Q_L)$ satisfies an LDP 
on $\CS^+_D$
with rate function
\[
I^*(Q)=\inf\{ I(Q_1,\dots,Q_L) : 
Q^{(L)}(Q_1,\dots,Q_L)=Q\}
\]
\end{remark}

After the re-scaling in Eq. \eqref{GLikeFL}, 
 the posterior predictive 
distribution is  the same mixture of Gaussians
described in 
Proposition~\ref{Prop2} with the posterior 
mixing measure  $\mathcal{Q}_{L,N}(\cdot| \by_{1:P},\beta)$   replaced by 
the rescaled mixing measure 
\[
 \mathcal{Q}_{L,N}^\circ(dQ_1 \dots dQ_L | \by_{1:P},\beta)   :=
 \frac{e^{-\frac{N}{2} \Phi^\circ_\beta(Q_1,\dots,Q_L,\by_{1:P}) 
 -\frac{1}{2} R(Q_1,\dots,Q_L)
 }\mathcal{Q}_{L,N}(Q_1,\dots,Q_L)  }{\int_{(\CS^+_{D})^L}e^{-\frac{N}{2}\Phi^\circ_\beta(Q_1,\dots,Q_L,\by_{1:P}) 
 -\frac{1}{2} R(Q_1,\dots,Q_L) } \mathcal{Q}_{L,N}(Q_1,\dots,Q_L), }  
\]
where
\begin{equation}\label{defPhi0}
\begin{split}
& \Phi_\beta^\circ(Q_1,\dots,Q_L,\by_{1:P}):=
\by_{1:P}^\top ( \Sigma_{11} +\beta^{-1}  \eI_{DP})^{-1} \by_{1:P} \\
& R(Q_1,\dots,Q_L)=\log(\det(\eI_{DP} +\beta \Sigma_{11})), \\
\end{split}
\end{equation}
and $\mathbf{m}_0$
replaced by  
 $\sqrt{N}\mathbf{m}_0$.
Combining well-known 
Varadhan's lemma together with
Proposition~\ref{LDPeasy},  one can prove that 
$\mathcal{Q}_{L,N}^\circ(dQ_1 \dots dQ_L | \by_{1:P},\beta)$ also satisfies an LDP. 

\begin{proposition}\label{LDP_postVara}
The sequence of measures $\mathcal{Q}^{\circ}_{L,N}(dQ_1,\dots,Q_L| \by_{1:P},\beta)$   as $N \to +\infty$ 
satisfies an LDP with rate function 
\[
I^\circ(Q_1,\dots,Q_L)=
\frac{1}{2}
\sum_{\ell=1}^L \Big(  \tr(Q_\ell)-
\log(\det(Q_\ell ))\Big)
+ \frac{1}{2} \Phi_\beta^\circ(Q_1,\dots,Q_L,\by_{1:P})
-\bar I_0.
\]
where $\bar I_0=\frac{1}{2} \inf_{Q_1,\dots,Q_L}\{ 
\sum_{\ell=1}^L \Big(  \tr(Q_\ell)-
\log(\det(Q_\ell))\Big)
+\Phi_\beta^\circ(Q_1,\dots,Q_L,\by_{1:P})\}$.
\end{proposition}

 The rate function now explicitly depends on the training inputs and labels, showing that the measure over the Wishart ensembles this time concentrates around non-trivial data-dependent solutions, which quantitatively characterize feature learning in this limit (see also Discussion in Section~\ref{sec3} for a comparison with the proportional limit studied in \citealp{SompolinskyLinear, pacelli2023statistical, doi:10.1073/pnas.2301345120}). We observe that the result in Proposition~\ref{LDP_postVara} is complementary to the recent investigation in  \cite{https://doi.org/10.1002/cpa.22200}, where the feature learning infinite-width limit of DLNs under gradient descent dynamics was characterized. 

\begin{remark}
{\rm Propositions~\ref{LDPeasy}}  and {\rm~\ref{LDP_postVara}} 
 hold  for the C-DLN with 
$\mathcal{Q}_{L,N}$ and $\mathcal{Q}^\circ_{L,N}$
 replaced by  $\mathcal{Q}_{L,C}$ and
$\mathcal{Q}^\circ_{L,C}$. In this case 
$C \to +\infty$ and 
$\Phi_\beta^\circ$ has the same form 
given in \eqref{defPhi0}
but $\Sigma_{11}$ is defined as in Proposition~\ref{Prop2-CNN}.
 \end{remark}

\section{Discussion and conclusions\label{sec:discussion}}

In this paper, we derived exact non-asymptotic integral representations of the prior distribution over the output and of the predictive posterior of deep linear Bayesian NNs, both for the case of fully-connected architectures with multiple outputs and of convolutional architectures. In this section, we provide a thorough discussion on how our results map to the ones already obtained in the past for special cases, as reported in Sec.~\ref{sec:related}, putting our work in perspective with the existing literature.

\subsection{Connection with the formalism of Meijer G-functions for $D=1$}\label{Meijer-Appendix}

In this section we provide a way to map our results for fully-connected NNs with scalar output ($D=1$) to the formalism of Meijer G-functions (see, for example, \citet{zavatone-veth2021exact,doi:10.1073/pnas.2301345120}). 
The most direct way to obtain this mapping is to consider
the Bayesian model evidence in the limit $\beta \to +\infty$
(zero temperature). 
With our notation (see Eq. \eqref{normalization_Z}) and recalling that here $D=1$
\[
\begin{split}
Z_{\by_{1:P},\beta}
& =\int_{\RE^{P+1}} \CL( \by_{1:P}|\bs_{1:P})
p_{\mathrm{prior}}(\bs_0,\bs_{1:P}|\bx^0,\bX)d\bs_{0} d\bs_{1:P} \\
&=\int_{\RE^{P}} \CL( \by_{1:P}|\bs_{1:P})
p_{\mathrm{prior}}(\bs_{1:P}|\bX) d\bs_{1:P} \\
&=\int_{(\RE^+)^L}e^{-\frac{1}{2} \Phi_\beta(Q_1 \dots Q_L,\by_{1:P}) } \mathcal{Q}_{L,N}(dQ_1 \dots dQ_L)
\end{split}
\]
with
 \[
\mathcal{Q}_{L,N} (d  Q_1 \dots d  Q_L)
= \prod_{\ell=1}^L
\frac{(N_\ell/2)^{N_\ell/2}  }{\Gamma(N_\ell/2)}   
(Q_\ell)^{\frac{N_\ell}{2}-1} e^{- \frac{N_\ell Q_\ell}{2} }  dQ_\ell .
\]
Taking the limit  $\beta\to\infty$
one obtains 
\[
Z_\infty:=
\lim_{\beta \to +\infty} \beta^{-P/2}Z_{\by_{1:P},\beta}
=
\int_{(\mathbb{R}^+)^L}e^{-\frac{ \by_{1:P}^\top \tilde{\Sigma}_{11}^{-1} \by_{1:P} }{2 Q_1 \dots, Q_L}  }  ( Q_1 \dots, Q_L)^{-P/2} \mathcal{Q}_{L,N}(dQ_1 \dots dQ_L) 
\]
where 
 $
\tilde{\Sigma}_{11} :=\frac{\bX^\top \bX}{N_0 \lambda^*}$. 
The resulting  $Z_\infty$ is (up to a constant) the Bayesian evidence  $Z_\infty(0)$  defined  in equation  [8]  of 
\citet{doi:10.1073/pnas.2301345120}. 
Recalling 
\eqref{dens_Meijer}, 
 the above integral becomes 
\[
Z_\infty = \frac{(\prod_{\ell=1}^L N_\ell/2)^{P/2}}{\prod_{\ell = 1}^L \Gamma(N_\ell/2) } \int_{\mathbb{R}^+}e^{-\frac{ \omega }{q  } } {q}^{-P/2}   G^{L0}_{0L}\!\left(q \mid \begin{matrix} - \\ \frac{N_1 }{2} - 1, \cdots, \frac{N_L}{2} - 1  \end{matrix}\right) dq \,,
\]
where
\[
\omega := \by_{1:P}^\top \tilde{\Sigma}_{11}^{-1} \by_{1:P}  \frac{\prod_{\ell=1}^L N_\ell }{2^{L+1}} = \by_{1:P}^\top (XX^\top)^{-1} \by_{1:P} \frac{ \prod_{\ell=0}^L \lambda_\ell N_\ell }{2^{L+1}}\,.
\]
With the change of variables $u = q^{-1}$, this integral becomes
\[
\begin{aligned}
Z_\infty &= \frac{(\prod_{\ell=1}^L N_\ell/2)^{P/2}}{\prod_{\ell = 1}^L \Gamma(N_\ell/2) } \int_{\mathbb{R}^+}e^{- u\omega  } u^{P/2-2}   G^{L0}_{0L}\!\left(u^{-1} \mid \begin{matrix} - \\ \frac{N_1 }{2} - 1, \cdots, \frac{N_L}{2} - 1  \end{matrix}\right) du \\
& = \frac{(\prod_{\ell=1}^L N_\ell/2)^{P/2}}{\prod_{\ell = 1}^L \Gamma(N_\ell/2) } \int_{\mathbb{R}^+}e^{- u\omega  } u^{P/2-2}   G^{0L}_{L0}\!\left(u \mid \begin{matrix} 2- \frac{N_1 }{2}, \cdots, 2-\frac{N_L}{2} \\ -  \end{matrix}\right) du \\
& = \frac{(\prod_{\ell=1}^L N_\ell/2)^{P/2}}{\prod_{\ell = 1}^L \Gamma(N_\ell/2) }  \omega^{1-P/2} G^{0,L+1}_{L+1,0}\!\left(\omega^{-1} \mid \begin{matrix} 2-\frac{P}{2},2- \frac{N_1 }{2}, \cdots, 2-\frac{N_L}{2} \\ -  \end{matrix}\right) \\
& = \frac{(\prod_{\ell=1}^L N_\ell/2)^{P/2}}{\prod_{\ell = 1}^L \Gamma(N_\ell/2) }  \omega^{1-P/2} G^{L+1,0}_{0,L+1}\!\left(\omega \mid \begin{matrix} - \\ \frac{P}{2} - 1,\frac{N_1 }{2}-1, \cdots, \frac{N_L}{2}-1 \end{matrix}\right)\\
& = \frac{(\prod_{\ell=1}^L N_\ell/2)^{P/2}}{\prod_{\ell = 1}^L \Gamma(N_\ell/2) }  \omega^{-P/2} G^{L+1,0}_{0,L+1}\!\left(\omega \mid \begin{matrix} - \\ \frac{P}{2},\frac{N_1 }{2}, \cdots, \frac{N_L}{2} \end{matrix}\right)\,,
\end{aligned}
\]
where we used at each step known properties of the Meijer G-functions (see Supplementary Material of~\citet{doi:10.1073/pnas.2301345120}, Eq. (7, 8, 9)). 
In this way one recovers  
expression [14] of Theorem 1 in 
\citet{doi:10.1073/pnas.2301345120}.
Along these lines, one can re-obtain also the other  results 
stated in Theorem 1 of ~\citet{doi:10.1073/pnas.2301345120} from our approach.

Moreover, 
Proposition~\ref{prop1} shows that $\bS_{1:P}$ is a mixture of 
Gaussians with a Wishart matrix as covariance. It is known that the resulting distribution is the so-called multivariate Student's $t$-distribution (see, e.g.,  Representation B in \citealp {Lin1972}). One can show that the characteristic function of a  multivariate Student's $t$ can be expressed in term of Macdonald functions (\citealp{Joarder1995}). Since  the Macdonald function is recovered as a special case of a Meijer G-function, this further clarifies (at least for $L=1$) the link between our stochastic representation and the Fourier representation given in  \cite{zavatone-veth2021exact}.

\subsection{From kernel renormalization to rigorous Bayesian statistics \label{sec3}} 

In concluding, we discuss how our derivations provide a way to interpret the recent statistical-physics literature on Bayesian deep learning theory in the proportional limit ($P, N_\ell \to \infty$, fixed $\alpha_\ell = P/N_\ell$) such as \cite{SompolinskyLinear, li2022globally, pacelli2023statistical, aiudi2023}.

\cite{doi:10.1073/pnas.2301345120} provides the first notable results that partly address this issue. However, the
representation of the Bayesian model evidence (partition function) in terms of Meijer G-functions hides the possibility of a direct comparison with the previously-established physics results.
On the contrary, our explicit non-asymptotic integral representation of the prior delivers a precise one-to-one mapping between the idea of kernel renormalization introduced by physicists (\citealp{SompolinskyLinear, pacelli2023statistical, aiudi2023}) and the Wishart ensembles that appear in our rigorous derivation.   

Let us first consider the case of single output DLNs, $D = 1$. Using the explicit integral representation for the prior given in Proposition~\ref{prop1}, and assuming Gaussian likelihood, we find the following non-asymptotic result for the partition function:
\begin{equation}
    Z_{\by_{1:P},\beta} \propto \int_0^\infty \prod_{\ell=1}^L dQ_\ell\, e^{-\sum_{\ell=1}^L\left[\frac{N_\ell}{2} Q_\ell -\left(\frac{N_\ell}{2}-1\right)\log Q_\ell\right] -\frac{1}{2} \Phi_\beta (Q_1, \dots, Q_L, \mathbf{y}_{1:P}) }\,,
    \label{pfsingle}
\end{equation}
which is exact up to an overall unimportant constant.
Note that here the first two terms in the exponential arise from the probability measure over the scalar variables $Q_\ell$, while the third and fourth data-dependent terms are a consequence of the Gaussian integration over the output variables. 
This partition function can be compared with the result for non-linear networks found in the proportional limit in \cite{pacelli2023statistical}.
It turns out that if we restrict our focus to linear NNGP kernels in \cite{pacelli2023statistical}, we find an effective action equivalent to the exponent in Eq.~\eqref{pfsingle} at leading order in $N_\ell$.
Moreover, it is 
reasonable to expect that the third term in Eq.~\eqref{pfsingle} scale linearly with $P$, under 
proper hypotheses on the training data (note that this requirement is similar to the one in \citealp{doi:10.1073/pnas.2301345120}). This
observation suggests that it is 
legitimate to employ large deviations techniques (or the more powerful saddle-point method) in the proportional limit, and it 
provides the interpretation mentioned
at the beginning of this section, since one can prove that the saddle-point equations derived from Eq.~\eqref{pfsingle} are exactly the same as those found in \cite{SompolinskyLinear}. Let us check this explicitly in the case $N_\ell = N$, $\forall \ell = 1, \dots, L$. The leading order of the exponent (at large $N$ and $P$) in Eq. \eqref{pfsingle} is given by: 
\begin{equation}
 -\frac{N}{2}\left\{\sum_{\ell=1}^L \left[Q_\ell - \log Q_\ell\right] + \frac{\alpha}{P}\by_{1:P}^\top ( \Sigma_{11} +\beta^{-1}  \eI_{P})^{-1} \by_{1:P}
 +\frac{\alpha}{P}\log(\det( \eI_{P} +\beta \Sigma_{11}))\right\}\,,
 \label{efaction}
\end{equation}
where $\alpha = P/N$ and $\Sigma_{11}$ is the same as defined in Proposition \ref{Prop2}. Assuming that the third and fourth terms of the equation above converge to a finite value in the proportional limit ($P,\,N \to \infty$ with $\alpha = P/N$ fixed), we can employ the saddle-point method to estimate the integral in Eq. \eqref{pfsingle}, which amounts to find the minima of the term in brackets in Eq. \eqref{efaction}. After derivation wrt each $Q_\ell$ and by noting that the only solution of the resulting system of $L$ equations has $Q_\ell = u_0$ $\forall \ell = 1,\dots, L$, we recover Eq. (11) in \cite{SompolinskyLinear} if we take the limit $\beta \to \infty$.

By a line of reasoning similar to that exposed above for $D = 1$, we can also interpret the (heuristic) kernel shape renormalization pointed out in \cite{SompolinskyLinear} 
and \cite{pacelli2023statistical} for one-hidden-layer fully-connected architectures with multiple outputs, as well as the local kernel renormalization proposed to describe shallow CNNs in \cite{aiudi2023}. If we consider for instance the effective action derived in ~\cite{aiudi2023}, we again recognize that the first two terms arise from the probability measure of a Wishart ensemble, whereas the last two data-dependent terms are a consequence of the integration of the output variables. Even in this case, one can check a precise equivalence with the non-asymptotic partition function that one finds in the CNN case using our result for the prior in Proposition~\ref{Prop2} for $L = 1$.

Finally, we point out that the feature learning infinite-width limit investigated in Proposition~\ref{LDP_postVara} shares qualitative similarities with the proportional limit studied in the physics literature. In both cases, the measure over the Wishart ensembles concentrates over data-dependent solutions of a rate function (or a saddle-point effective action).  However, this happens for different reasons in the two settings: in the feature learning infinite-width limit, this effect is due to the additional data-dependent terms entering the rate function thanks to the mean-field scaling; in the proportional limit with standard scaling discussed in this section, data are instead entering the saddle point equations for the variables $Q_\ell$ in the action~\eqref{efaction}.
Furthermore, the three terms of the rate function in Proposition~\ref{LDP_postVara} also appear in the effective action found in \cite{pacelli2023statistical}, which is known to reproduce the saddle-point equations firstly identified in \cite{SompolinskyLinear} for the linear case.


\acks{P.R. is supported by $\#$NEXTGENERATIONEU (NGEU) and funded by the Ministry of University and Research (MUR), National Recovery and Resilience Plan (NRRP), project MNESYS (PE0000006) ``A Multiscale integrated approach to the study of the nervous system in health and disease'' (DN. 1553 11.10.2022). F.B. is partially supported by the MUR - PRIN project ``Discrete random structures for Bayesian learning and prediction'' no. 2022CLTYP4.}


\newpage

\appendix
\section{Proofs}

\subsection{Preliminary facts}
 {\it Matrix normal distribution.} A random matrix 
$Z$ of dimension $n_1 \times n_2$  has a centred matrix normal distribution 
with parameters $(\Sigma_1,\Sigma_2)$ (with $\Sigma_i$'s positive symmetric $n_i \times n_i$ matrices),
if for any  matrix $S$ of dimension $n_2 \times n_1$ 
\begin{equation}\label{PMN1}
\E[e^{i \tr(S Z)}]=\exp\left \{-\frac{1}{2} \tr(S\Sigma_1 S^\top \Sigma_2) \right\}.
\end{equation}
In symbols $Z \sim \CMN(0,\Sigma_1,\Sigma_2)$. 

We collect here some properties of multivariate normals, matrix normals
and Wishart distributions to be used later.  
See \cite{Gupta2000} for details. 
In what follows $\vvec(A)$ denotes the stacking of the {\it columns} of a matrix $A$ to form a vector.

\begin{enumerate}[label={(P\arabic*)},start=1]
 \label{P1}
\item {\it  Linear transformation of matrix normals. }
 Given two matrices $H$ and $K$ with compatible shape, if $Z \sim \CMN(0,\Sigma_1,\Sigma_2)$
then 
\[
H Z K \sim \CMN(0,H\Sigma_1H^\top,K^\top\Sigma_2K).
\]
 \label{PMN2}
 \item
{\it    Equivalence with the multivariate normal. }
  $Z \sim \CMN(0,\Sigma_1,\Sigma_2)$ if and only if 
$\vvec(Z) \sim \CN(0,\Sigma_2\otimes \Sigma_1)$
  \label{PMN3}
  \item {\it  Laplace functional of a Wishart distribution}.
   \label{P4}
Let $S$ be a symmetric positive definite matrix. Then 
  for any  symmetric  matrix $C$  and any real number $\alpha$ such 
  that all the eigenvalues of 
  $(\eI_{D}+ \alpha S C  )$ are strictly positive,   
  \begin{equation}\label{LaplaceWishart}
 \begin{split}
 [\det(\eI_{D}+ \alpha S C )]^{-\frac{N}{2}}
&  = \int_{\CS^+_D } e^{-\frac{\alpha}{2} \tr(CQ )} 
\CW_D (dQ|S,N).
\\
\end{split}
   \end{equation}
   Versions of this formula can be found in literature as \emph{Ingham-Siegel integrals}~(\citealp{ingham1933,siegel1935}, and ~\citealp{fyodorov2002} for an extension to Hermitian matrices); in this generality, the statement can be derived by Proposition 8.3  and its proof in \cite{Eaton2007}. 
 In particular the formula above holds 
 for $\alpha>0$ and $S=\eI_D$ and $C$
 symmetric and $\geq 0$, showing that 
when $\mathbf{Z}_j  \stackrel{iid}{\sim} \CN(0,C)$, then
 \begin{equation}\label{P4newbis}
      \E[e^{-\frac{\alpha}{2} \sum_{j=1}^N  \|\mathbf{Z}_j \|^2} ]=
   \int_{\CS^+_D } e^{-\frac{\alpha}{2} \tr(CQ )} 
\CW_D (dQ|\eI_D,N).
   \end{equation}
   \label{P4NEW} 
\end{enumerate}

\subsection{Proofs for the FC-DLN}

We set   
$h_{i_\ell,\mu}^{(\ell)}=h_{i_\ell}^{(\ell)}(\bx^{\mu})$
and
\[
H^{0}:=\bX =[\bx^{1},\dots,\bx^{P}].
\]
For $\ell \geq 2$ define the   $N_{\ell-1} \times P$   matrix 
\[
H^{(\ell-1)}=[ h_{i,\mu}^{(\ell-1)}]_{i,\mu}. 
\]
Notice that the weights of the network distributed as~\eqref{lawofW} can be arranged as rectangular matrices with law
\[
 W^{(\ell-1)} \sim \CMN(0,\eI_{N_{\ell}},{\lambda_{\ell-1}^{-1}} \eI_{N_{\ell-1}}).
\]
With this notations, it is immediate to check that  
for any $\ell$ one has 
\begin{equation}\label{FCLrecursion0}
H^{(\ell)}=\frac{1}{\sqrt{N_{\ell-1}}} W^{(\ell-1)} H^{(\ell-1)}\,.
\end{equation}

\begin{enumerate}[label={(P\arabic*)},start=4]
\item Let $\mathbf{H}^{(\ell)}_{i_{\ell}}
=(h_{i_{\ell},1}^{(\ell-1)},\dots,h_{i_{\ell},P}^{(\ell-1)})^\top$ the $i_\ell$-th row of 
$H^{(\ell)}$. Given $H^{(\ell-1)}$, the vectors $\mathbf{H}^{(\ell)}_{i_{\ell}}$ 
are independent with law  $\CN(0, (\lambda_{\ell-1} N_{\ell-1})^{-1} H^{(\ell-1)\top}H^{(\ell-1)})$.
Indeed,  $\mathbf{H}^{(\ell)}_{i_{\ell}}=N_{\ell-1}^{-\frac{1}{2}} W^{(\ell-1)}_{i_{\ell},\cdot} H^{(\ell-1)}  $ and the $W^{(\ell-1)}_{i_{\ell},\cdot}$'s are independent and identically distributed.
Recall that if $\mathbf{Z} \sim \CN(0,C)$ 
and $A$ is a matrix 
then $ A\mathbf{Z}  \sim \CN(0,ACA^\top)$.
\label{Step0}
\end{enumerate}

\begin{proof}{\bf~of \eqref{basic_rep_in_matrix_form} in Remark~\ref{Rem2}}  
Iterating \eqref{FCLrecursion0} one gets  
\begin{equation}\label{FCLrecursion1}
H^{(L+1)}= \frac{1}{\sqrt{N_{L}}} W^{(L)} H^{(L)}   =\cdots= 
\frac{1}{\sqrt{N_{L}}}  W^{(L)} \cdots 
\frac{1}{\sqrt{N_{0}}}  W^{(0)} \bX\,.
\end{equation}
\end{proof}

\begin{proof}{\bf~of Proposition~\ref{prop1} and  of  \eqref{main_in_matrix_form} in Remark~\ref{Rem2}} 
Starting from \eqref{FCLrecursion0} with $\ell=L+1$, 
conditioning on $H^{(L)}$, by~\ref{PMN2} and \eqref{PMN1}, it follows that for any $P \times N_{L+1} = P\times D $ 
 matrix  $\bar S$ 
 \begin{equation}\label{Step1}
\E[e^{i \tr(H^{(L+1)}\bar S )}]= \E[e^{-\frac{1}{2N_{L} \lambda_L} \tr(\bar S^\top H^{(L)\top} H^{(L)} \bar S )}] \,.
\end{equation}
We divide the proof in few steps. 

 {\bf Step 1.} For any $\ell \geq 1$ and any $P\times D$ 
 matrix  $\Theta$ 
\begin{equation}\label{step2}
\tr(\Theta^\top H^{(\ell)\top} H^{(\ell)}  \Theta )
=\sum_{i_{\ell}=1}^{N_{\ell}} \| \mathbf{Y}_{i_{\ell}}^\ell\|^2 
\end{equation}
with $\mathbf{Y}_{i_{\ell-1}}^\ell=  H^{(\ell)}_{i_{\ell},\cdot}\Theta $.
To see this, note that 
\[
\begin{split}
\tr(\Theta^\top H^{(\ell)\top} H^{(\ell)}  \Theta )
& =\sum_{i} \sum_{\mu,\nu} \Theta_{\mu,i}  \Big( \sum_{i_{\ell}} h^{(\ell)}_{i_{\ell},\mu} h^{(\ell)}_{i_{\ell},\nu} \Big) \Theta_{\nu, i}  =\sum_{i_{\ell}} 
\sum_{i} \sum_{\mu,\nu}\Theta_{\mu,i}  \Theta_{\nu,i}
h^{(\ell)}_{i_{\ell},\mu} h^{(\ell)}_{i_{\ell},\nu} 
\\
& 
= \sum_{i_{\ell}} 
\sum_{i}  \Big( \sum_{\mu} h^{(\ell)}_{i_{\ell},\mu} \Theta_{ \mu,i}
 \Big)^2
=\sum_{i_{\ell}} \|\mathbf{Y}_{i_{\ell}}^\ell\|^2 .
\\
\end{split}
\]

{\bf Step 2.}  
For any $\ell \geq 1$ and any $P \times D$ matrix $\Theta$
we claim that 
\begin{equation}\label{step3}
\E[e^{-\frac{1}{2 \lambda_\ell N_\ell} \tr(\Theta^\top H^{(\ell)\top} H^{(\ell)} \Theta )}]=
\E[ e^{-\frac{1}{2\lambda_{\ell} } \tr(  K_{\ell-1} Q_{\ell})} ]
\end{equation}
where $Q_{\ell}$ is a $D \times D$  random matrix  with  Wishart distribution
 of  $N_{\ell}$ degree of freedom and
 scale matrix $\eI_{D}/N_\ell$ and  
 \[
 K_{\ell-1} :=\frac{1}{\lambda_{\ell-1} N_{\ell-1}}  \Theta^\top H^{(\ell-1)\top} H^{(\ell-1)} \Theta .
 \]
 Writing $Q_{\ell}=U_{\ell} ^\top U_{\ell}$, with $U_\ell$ a $D \times D$ matrix, one has
\begin{equation}\label{to-iterate}
\E[e^{-\frac{1}{2 \lambda_\ell N_\ell} \tr(\Theta^\top H^{(\ell)\top} H^{(\ell)}  \Theta )}]
= \E[ e^{- \frac{1}{2 \lambda_{\ell} N_{\ell-1}
\lambda_{\ell-1} }   \tr(U_{\ell}   \Theta^\top H^{(\ell-1)\top} H^{(\ell-1)}\Theta U_{\ell-1} ^\top) } ].
\end{equation}
To prove \eqref{step3},  note that 
by~\ref{Step0}
the
 $\mathbf{Y}_{i_{\ell}}^\ell$'s are independent and identically distributed conditionally on $H^{\ell-1}$ , with 
\[
\mathbf{Y}_{i_{\ell}}^\ell | H^{(\ell-1)} \sim  \CN(0, K_{\ell-1}  ).
\]
Hence using, \eqref{step2} and \eqref{P4newbis}
\begin{equation}
 \E[e^{-\frac{1}{2 \lambda_{\ell}N_{\ell} } \tr(\Theta^\top H^{(\ell)\top} H^{(\ell)}  \Theta )}]
= \E[e^{-\frac{1}{2 \lambda_{\ell}N_{\ell}} \sum_{i_{\ell}=1}^{N_{\ell}} \| \mathbf{Y}_{i_{\ell}}^\ell\|^2  }]
=\E[ e^{-\frac{1 }{2 \lambda_{\ell} } \tr(  K_{\ell-1} Q_{\ell})} ].
\end{equation}

{\bf Step 3.}   
 Starting from $\ell=L$ and  $\Theta=\bar S$, 
 iterating \eqref{to-iterate} one obtains 
\[
\E[e^{-\frac{1}{2\lambda_{L}N_L} \tr(\bar S^\top H^{(L)\top} H^{(L)}  \bar S )}]
= \E[ e^{- \frac{1}{2 \lambda^* N_0}
\tr(U_{1:L}  \bar S^\top H^{(0)\top} H^{(0)} \bar S U_{1:L}^\top) } ]
\]
where $\lambda^*:=\lambda_0 \dots \lambda_L$
and $U_{1:L}=U_1 \dots U_{L}$  for 
 $Q_\ell=U_{\ell}^\top U_{\ell}$
 independent $D\times D$ random matrix with Wishart distribution 
 with $N_{\ell}$ degree of freedom
 and $\eI_{D}/N_\ell$ as scale matrix. 
By \eqref{Step1} we conclude that 
\[
\begin{split}
\E[e^{i \tr(H^{(L+1)}\bar S)}] & =
\E[e^{-\frac{1}{2 \lambda_{L} N_L} \tr(\bar S^\top H^{(L)\top} H^{(L)}  \bar S )}] \\
& 
= \E[ e^{- \frac{1}{2\lambda^* N_0}  \tr(U_{1:L}  \bar S^\top H^{(0)\top} H^{(0)} \bar S U_{1:L}^\top) } ]
\\
&= \E[ e^{- \frac{1}{2\lambda^* N_0} \tr(
\bar S  U_{1:L}^\top U_{1:L}
\bar S^\top H^{(0)\top} H^{(0)}  ) } ]\,.
\end{split}
\]
Recalling that 
$H^{(0)\top} H^{(0)} :=\bX^\top\bX$, 
using once again \eqref{PMN1}, this shows that, conditionally on 
$U_{1:L}^\top U_{1:L}$, 
$H^{(L+1)} \sim \CMN(0,U_{1:L}^\top U_{1:L},(\lambda^*N_0)^{-1}\bX^\top \bX)$.
Note that this is 
\eqref{main_in_matrix_form}
of Remark~\ref{Rem2}. 

Since $\bS_{1:P}=  \vvec[H^{(L+1)}]$,
by~\ref{PMN3}, 
one gets
\[
\bS_{1:P} | U_{1:L}^\top U_{1:L} 
\sim \CN(0,  (\lambda^*N_0)^{-1} \bX^\top \bX \otimes U_{1:L}^\top U_{1:L} ).
\]
and the thesis follows. 
\end{proof}

We now prove a slightly more general statement of the one given in Proposition~\ref{Prop2}.
Let  $\bs^\top=((\bs^0)^\top,(\bs^1)^\top,\dots,(\bs^P)^\top)$ 
and $\tilde \bX=[\bx^0,\bx^1,\dots,\bx^P]$.
In what follows $\eI_{M}$ is the identity matrix 
of dimension $M \times M$ and $0_{M \times N}$
is the zero matrix of dimension $M \times N$.

\begin{proposition}\label{posterior_joint}
Let 
$\mathcal{Q}_{L,N}(dQ_1 \dots dQ_L)$ 
be as in Eq. \eqref{def:QLN} 
and $Q^{(L)}=Q^{(L)}(Q_1,\dots,Q_L)$ 
be defined as in {\rm Proposition~\ref{prop1}}.
Let  $p_{\mathrm{post}}(\bs|\by_{1:P},\bx_0,\bX)$
be the posterior distribution 
 of  
$(\bS_0,\dots,\bS_P)$ given $\by_{1:P}$ in a 
 FC-DLN.
If
 $\det(\tilde \bX^\top \tilde\bX)>0$, then  
\[
    p_{\mathrm{post}}(\bs |\by_{1:P},\tilde \bX) 
= \int_{(\CS^+_{D})^L} \frac{
e^{- \frac{1}{2}(\bs- \mathbf{m})^\top  (\beta \Pi_0 +\Sigma^{-1}) (\bs-\mathbf{m}) }
}{(2 \pi)^{\frac{D(P+1)}{2}}\det(  
(\beta \Pi_0 +\Sigma^{-1})^{-1}
)^{\frac{1}{2}}  }
 \mathcal{Q}_{L,N}(dQ_1\dots dQ_L| \by_{1:P},\tilde \bX,\beta) 
 \]
where  
\[
\Sigma=\begin{pmatrix}
\Sigma_{00} &  \Sigma_{01} \\
\Sigma_{01}^\top &  \Sigma_{11} \\
\end{pmatrix}
:= \frac{\tilde \bX^\top \tilde \bX}{N_0 \lambda^*} \otimes   Q^{(L)},
\qquad
\Pi_0=
\begin{pmatrix}
0_{D \times D}  & 0_{ P \times DP}  \\
0_{DP\times P}  &  \eI_{DP} \\
\end{pmatrix}
\]
\[
 \mathbf{m}:=
 \begin{pmatrix}
      \mathbf{m}_0\\
      \mathbf{m}_1 \\
 \end{pmatrix}
=
 \begin{pmatrix}
      \Sigma_{01}  
 (  \Sigma_{11} +\beta^{-1} \eI_{DP})^{-1}  \\
  \Sigma_{11} ( \Sigma_{11} +\beta^{-1} \eI_{DP})^{-1}\\
 \end{pmatrix}\by_{1:P},
 \]
 \[
 \mathcal{Q}_{L,N}(dQ_1\dots dQ_L| \by_{1:P},\tilde \bX,\beta)   :=
 \frac{e^{-\frac{1}{2} \Phi_\beta(Q_1,\dots, Q_L,\by_{1:P}) }\mathcal{Q}_{L,N}(dQ_1\dots dQ_L) }{\int_{(\CS^+_{D})^L}e^{-\frac{1}{2} \Phi_\beta(Q_1,\dots, Q_L,\by_{1:P}) } \mathcal{Q}_{L,N}(dQ_1\dots dQ_L) }  
 \]
 and 
$\Phi_\beta(Q_1,\dots, Q_L,\by_{1:P}):=
\by_{1:P}^\top ( \Sigma_{11} +\beta^{-1}  \eI_{DK})^{-1} \by_{1:P}
 +\log(\det(\eI_{DP} +\beta \Sigma_{11}))
$. 
\end{proposition}

\begin{proof}
Proposition~\ref{prop1} gives 
\[
\CL( \by_{1:P}|\bs_{1:P})
p_{\mathrm{prior}}(\bs_0,\bs_{1:P}|\bx^0,\bX)=
\int\frac{ e^{-\frac{\beta}{2} (\bs_{1:P}-\by_{1:P})^\top(\bs_{1:P}-\by_{1:P}) 
-\frac{1}{2} \bs\Sigma^{-1} \bs }} 
{\det(\Sigma)^{\frac{1}{2} } (2\pi)^{\frac{D(P+1)}{2}}}  
\mathcal{Q}_{L,N}(dQ_1\dots dQ_L) 
\]
with  
 $\Sigma:= (N_0 \lambda^*)^{-1} \tilde\bX^\top \tilde \bX \otimes  Q^{(L)} $.
Since $\tilde \bX^\top \tilde \bX$ is assumed to be  strictly positive definite 
and hence invertible,  using that  $(A \otimes B)^{-1}
=A^{-1} \otimes B^{-1}$ (see, e.g., 
Proposition 1.28 \citealp{Eaton2007})
it follows 
that also $\Sigma$ is invertible. 
Simple computations show that 
\[
\begin{split}
{\beta}(\bs_{1:P}-\by_{1:P})^\top(\bs_{1:P}-\by_{1:P})+
\bs \Sigma^{-1} \bs & =
(\bs- \tilde{ \mathbf{y}})^\top {\Pi_0}{\beta}(\bs- \tilde{ \mathbf{y}})+
\bs \Sigma^{-1} \bs \\
& = (\bs-\mathbf{m})^\top ({\Pi_0}{\beta} +\Sigma^{-1}) (\bs- \mathbf{m}) + \varphi(\Sigma,\by_{1:P}) \\
\end{split}
\]
where 
\[
\begin{split}
& \tilde{ \mathbf{y}}:=(0_{D}^\top,\by_{1:P}^\top)^\top, \quad
 \mathbf{m}=({\Pi_0}{\beta} +\Sigma^{-1})^{-1}{\Pi_0}{\beta}  \tilde{ \mathbf{y}}
={\beta}({\Pi_0}{\beta} +\Sigma^{-1})^{-1}  \tilde{ \mathbf{y}}\,,
 \\
& \varphi(\Sigma,\by_{1:P})=
\beta{\by_{1:P}^\top \by_{1:P}}- (\mathbf{m})^\top ({\Pi_0}{\beta} +\Sigma^{-1})\mathbf{m} \,.\\
 \\
\end{split}
\]
Writing $\mathbf{m}:=[\mathbf{m}_0, \mathbf{m}_1]$ one can  check that   
 $( \Sigma_{11}^{-1} +\beta \eI_{DP})^{-1} 
 =\Sigma_{11}(\eI_{DP} +\beta \Sigma_{11})^{-1}$ and 
 \[
 \begin{split}
& \mathbf{m}_1= \beta( \Sigma_{11}^{-1} +\beta \eI_{DP})^{-1} \by_{1:P}=
\beta \Sigma_{11}(\eI_{DP} +\beta \Sigma_{11})^{-1}\by_{1:P},
\\
& \mathbf{m}_0=\beta \Sigma_{01} \Sigma_{11}^{-1} 
 ( \Sigma_{11}^{-1} +\beta \eI_{DP})^{-1} \by_{1:P}
 =\beta \Sigma_{01} (\eI_{DP} +\beta \Sigma_{11})^{-1} \by_{1:P},
 \\
 & 
 \varphi(\Sigma,\by_{1:P})=
\beta{\by_{1:P}^\top \by_{1:P}}-\beta^2 
 \by_{1:P}^\top ( \Sigma_{11}^{-1} +\beta \eI_{DP})^{-1} \by_{1:P}
 =\by_{1:P}^\top( \Sigma_{11}^{-1} + \eI_{DP}/\beta)^{-1} \by_{1:P}. \\
 \end{split}
\]
In conclusion 
\[
 \frac{e^{-\frac{\beta}{2} (\bs_{1:P}-\by_{1:P})^\top(\bs_{1:P}-\by_{1:P}) 
-\frac{1}{2} \bs \Sigma^{-1} \bs } }{ |\Sigma|^{\frac{1}{2} } }  =
 \frac{e^{-\frac{1}{2}\varphi(\Sigma,\by_{1:P})  }
   }{\det(\Sigma)^{\frac{1}{2}} \det(\beta{\Pi_0}+\Sigma^{-1} )^{\frac{1}{2} }}
 \frac{e^{- \frac{1}{2}(\bs- \mathbf{m})^\top( \beta \Pi_0+\Sigma^{-1} ) (\bs-\mathbf{m}) }}{
 (\det( \beta \Pi_0+\Sigma^{-1} )^{-1})^{\frac{1}{2}}  } .
\]
where $
\det(\Sigma)^{\frac{1}{2}} \det(\beta{\Pi_0}+\Sigma^{-1} )^{\frac{1}{2} }
=\det(\eI_{DP} +\beta \Sigma_{11})^{\frac{1}{2}}.
$
   
\end{proof}

\vskip 0.5cm
\begin{proof}{\bf of Proposition~\ref{Prop2}}
Applying the previous proposition, 
one gets  that  
\[
p_{\mathrm{pred}}(\bs_0|\by_{1:P},\tilde \bX)
=\int p_{\mathrm{post}}(\bs_0,\bs_1,\dots,\bs_P|\by_{1:P},\tilde \bX)d\bs_1 \dots d\bs_P
\]
is again 
a mixture  of gaussians with 
mean  $\mathbf{m}_0=
\Sigma_{01}  
 (  \Sigma_{11} +\beta^{-1} I)^{-1} \by_{1:P}$ 
and (conditional)  covariance 
\[
\Big[\Big (\beta \Pi_0 +\Sigma^{-1}\Big)^{-1}\Big]_{1:D,1:D}.
\]
To conclude it remains to compute the previous 
matrix.
We have already noted that, 
by $\tilde \bX^\top \tilde \bX>0$,
it follows 
that also $\Sigma$ is invertible. 
Being $\bX^\top \bX$ 
and $\bx_0^\top \bx_0$  
strictly positive, then 
the Schur complement 
of $\Sigma_{00}$ in $\Sigma$, 
$\Sigma^*_{00}:=\Sigma_{00}-\Sigma_{01}\Sigma_{11}^{-1} \Sigma_{01}^\top$, 
is well-defined and strictly positive
(Proposition 1.34 \citealp{Eaton2007}). 
In particular, 
using the Banachiewicz inversion formula, 
one can write the inverse of $\Sigma$ as 
\[
\Sigma^{-1}=\begin{pmatrix}
(\Sigma^*_{00})^{-1} &  
-(\Sigma^*_{00})^{-1} \Sigma_{01} \Sigma_{11}^{-1} \\
-((\Sigma^*_{00})^{-1} \Sigma_{01} \Sigma_{11}^{-1})^\top & 
\Sigma_{11}^{-1}+\Sigma_{11}^{-1}
\Sigma_{01}^\top(\Sigma^*_{00})^{-1} \Sigma_{01}
\Sigma_{11}^{-1}\\
\end{pmatrix}.
\]
See, e.g.,  Theorem 1.2 in \cite{Zhang2005}.
Using again the  Banachiewicz inversion formula
 on
$\beta \Pi_0 +\Sigma^{-1}$ (which is again strictly positive), after some tedious computations, one gets that 
\[
\Big[\Big (\beta \Pi_0 +\Sigma^{-1}\Big)^{-1}\Big]_{1:D,1:D}
=\Sigma_{00}- 
\Sigma_{01}\Big (\frac{1}{\beta}\eI_{DP} + \Sigma_{11}\Big )^{-1} \Sigma_{01}^\top. 
\]
Observing that 
\[
\Sigma_{00}=  \frac{\bx_0^\top \bx_0}{N_0 \lambda^*} \otimes  Q^{(L)}, \quad
\Sigma_{01}=\frac{\bx_0^\top \bX }{N_0 \lambda^*} \otimes Q^{(L)}  , \quad  
\Sigma_{11}=\frac{\bX^\top \bX }{N_0 \lambda^*} \otimes Q^{(L)} , 
\]
the proof of Proposition~\ref{Prop2}
is concluded.  
\end{proof}

\begin{proof}
{\bf of Proposition~\ref{LDPeasy}}
We give the proof in the case $L=1$,  the general case follows 
the same lines, or using Exercise 4.2.7 in \cite{DemboZeitouni}
and the independence of the components. The result is a consequence of a very general version of the Cram\'er's theorem. 
See Theorem 6.1.3  and Corollary 6.1.6 in \cite{DemboZeitouni}. 
Recall that if $Q_1 \sim \CW_D(\cdot|\eI_D,N)$
then 
$Q_\ell=\frac{1}{N} \sum_{j=1}^N \bZ_j \bZ_j^\top$
where $\bZ_j \bZ_j^\top \sim \CW_D(\cdot|\eI_D,1)$ are independent and identically distributed taking values in 
$\CS^+_D$. Now  $\CS^+_D$ is a convex  open cone
in the  (topological) vector space $\CS_D$ of the $D\times D$ symmetric matrices, endowed 
with the scalar product $(A,B)=\tr(AB)$. The space $\CS_D$ is locally convex, Hausdorff and 
the closure of $\CS^+_D$  in $\CS_D$ is convex and separable and complete (with the topology induced by $\CS_D)$. Clearly $P\{  \bZ_j \bZ_j^\top \in \overline{\CS^+_D}\}=1$ and 
any closed convex hull of a compact in $\overline{\CS^+_D}$ is compact. Hence 
all the hypotheses of the General Cram\'er's theorem given in Chapter 6 of \cite{DemboZeitouni}
are satisfied and hence  
$Q_\ell$ satisfies  a (weak) LDP  (when $N \to +\infty$) with rate function given by 
$
I(Q)=\Lambda^*(Q):=\sup_{A \in \CS_D }\{ \tr(QA)-\Lambda(A)\}$ where 
\[
\Lambda(A)=\log \left ( \int_{\CS^+_D} e^{ \tr(AQ)} P\{\bZ_1 \bZ_1^\top  \in dQ\} \right) .
\]
In point of fact, see Corollary 6.1.6 in  \cite{DemboZeitouni}, it holds also a strong LDP
provided that $0$ is in the interior 
of the domain of $\Lambda$, as we shall show. 
The well-known expression 
for the Laplace transform of a Wishart distribution with one degree of freedom (see \eqref{LaplaceWishart})
gives
$\Lambda(A)=-\frac{1}{2} \log(  \det(\eI_D -2A))$ 
wherever the eigenvalues of $\eI_D -2A$ are positive and $+\infty$ otherwise. 
In computing 
$\Lambda^*(Q)$ we can restrict 
$A$ to be such that the eigenvalues of $\eI_D -2A$ are positive, since otherwise  $-\Lambda(A)=-\infty$.
Now we check that 
\[
\Lambda^*(Q)=\sup_{A}\{ \tr(QA)+\frac{1}{2} \log \det(\eI_D -2A)\}
=\frac{1}{2}(\tr(Q)+ 
\log(\det(Q^{-1}))-D).
\]
Making the change of variables $B=\eI_D -2A$, we compute 
$\sup_B H(B,Q)$ for 
\[
H(B,Q)= \tr(Q(\eI_D-B)/2 )+\frac{1}{2} \log(\det(B))
\]
where now the $\sup$ is over symmetric  strictly positive matrices $B$.  
The gradient of $B \mapsto H(B,Q)$ is easily computed 
 with the 
Jacobi's formula, 
$\partial_{B_{ij}} \Lambda(A) = -Q_{ij}+ (B^{-1})_{ij}$ for $i<j$
and $\partial_{B_{ii}} \Lambda(A) = -Q_{ii}/2+ (B^{-1})_{ii}/2$ for $i=j$. 
Solving $\nabla H(B_*,Q)=0$ we get 
$0=-Q+B_*^{-1}$ and hence $B_*=Q^{-1}$. Since $B \mapsto -H(B,Q)$  is convex
and $-H(B,Q) \to +\infty$ when $B$ converges to  the boundary of the convex cone $\CS^+_D$
(since in this case $\log \det(B) \to -\infty$), the unique critical  point $B_*$ is also the unique minimum.
Hence $\Lambda^*(Q)=H(B_*,Q)=\frac{1}{2}(\tr(Q)+ 
\log(|Q^{-1}|)-D)$, as desired. 
The LDP for  $\mathcal{Q}_{L,N}(dQ_1 \dots dQ_L | \by_{1:P},\beta) $  follows now using the next  Lemma~\ref{Lem_Varadhan}. 
 Details are omitted 
 since very similar to those given 
 in the next proof of Proposition~\ref{LDP_postVara}. 
\end{proof}

\vskip 0.5cm 
In order to prove Proposition~\ref{LDP_postVara}, 
we need the following variant of the Varadhan's
lemma. 

\begin{lemma}\label{Lem_Varadhan}
Let ${P}_N$ be a sequence of probability measure 
on a convex closed subset $\mathcal{S} \subset \RE^D$, 
satisfying an LDP with rate function $I$.  
Assume that 
\begin{itemize}
\item $\Phi_0$ is a bounded continuous function 
from $\mathcal{S} \to \RE$;
\item $\rho :\mathcal{S} \to [0,+\infty)$ such that
$\sup_{O \cap B_R} |\rho(s)|<+\infty$ 
for any $R$, where $B_R=\{ s \in \RE^D :
\|s\| < R \}$;
\item the rate function $I$ is such that  
$\inf_{s \in  \mathcal{S} \cap B_R^c} 
I(s) \to +\infty$ as $R \to +\infty$ 
and $\inf_s I(s) <+\infty$. 
\end{itemize}
Then 
\[
P_N^{\circ}(ds)= \frac{e^{-(N\Phi_0(s)+\rho(s))} P_N(ds)}{
 \int_\mathcal{S}
e^{-(N\Phi_0(s) +\rho(s)) } P_N(ds)} 
\]
satisfies an LDP with rate function $I(s)+\Phi_0(s)-I_0$ where 
where $I_0=\inf_s [I(s)+\Phi_0(s)]$.
\end{lemma}

\begin{proof}
Varhadan Lemma gives that 
$\bar P_N(ds)=P_N(ds)/
(\int_\mathcal{S}
e^{-N \Phi_0(s)} P_N(ds))$
satifies an LDP with rate 
$I_{\Phi_0}(s)=I(s)+\Phi_0(s)-I_0$.
Then, 
 \[
 \frac{1}{N} \log\Big (\int_\mathcal{S}
e^{-(N \Phi_0(s)+\rho(s))} \bar P_N(ds)\Big) 
\leq  \frac{1}{N} \log\Big (\int_\mathcal{S}
e^{-N \Phi_0(s)} \bar P_N(ds) \Big) 
=0
 \]
and
\[
\begin{split}
\liminf_N \frac{1}{N} &  \log \Big(\int_\mathcal{S}
e^{-(N \Phi_0(s)+\rho(s))} \bar P_N(ds) \Big )
\geq \liminf_N 
\frac{1}{N} \log \Big(
\int_{B_R \cap \mathcal{S} }
e^{-N \Phi_0(s)} \bar P_N(ds)  e^{-\sup_{B_R \cap \mathcal{S}} |\rho(s)|} \Big )  \\
& 
\geq -\inf_{s \in B_R \cap \mathcal{S} }
[I(s)+\Phi_0(s)]-I_0 . \\
\end{split}
\]
Since for $R \to +\infty$ one has  $\inf_{s \in B_R \cap \mathcal{S} }
[I(s)+\Phi_0(s)]-I_0 \to 0$, one obtains
that 
\[
\frac{1}{N}   \log \Big(\int_\mathcal{S}
e^{-(N \Phi_0(s)+\rho(s))} \bar P_N(ds) \Big ) \to 0
\]
which means that 
\[
\frac{1}{N}  \log \Big(\frac{\int_\mathcal{S}
e^{-(N \Phi_0(s)+\rho(s))}  P_N(ds)}{
\int_\mathcal{S}
e^{-N\Phi_0(s)}  P_N(ds)
} \Big ) \to 0.
\]
If now $C$ is a closed set, then 
\[
\begin{split}
& \limsup_{N \to +\infty} \frac{1}{N} \log \Big(  \int_C 
e^{-(N \Phi_0(s)+\rho(s))]} \bar P_N(ds) \Big)    \leq \limsup_{N \to +\infty} 
 \frac{1}{N} \log \Big( \int_C 
e^{-N \Phi_0(s)} \bar P_N(ds) \Big )\\
& \leq-\inf_{s \in C}
[I(s)-\Phi_0(s)-I_0] \\
\end{split}
\]
On the other hand if $O$ is open, then 
\[
\begin{split}
& \liminf_{N \to +\infty}  \frac{1}{N} \log \Big( \int_O
e^{-(N \Phi_0(s)+\rho(s))]} \bar P_N(ds)
 \log \Big)  \\
 &\geq 
\liminf_{N \to +\infty}\frac{1}{N}
\Big [ \log \Big(\int_{O \cap B_R} 
e^{-N \Phi_0(s)} \bar P_N(ds)
\log \Big)
+\log \Big(e^{-\sup_{O \cap B_R} |\rho(s)|} \Big)
\Big]\\
& \geq -\inf_{s \in O \cap B_R}
[I(s)+\Phi_0(s)-I_0]. \\
\end{split}
\]
Since 
$\inf_{s \in  \mathcal{S} \cap B_R^c} 
I(s) \to +\infty$ as $R \to +\infty$, $\inf_s I(s) <+\infty$ and $\Phi_0$
is bounded 
$\inf_{s \in  \mathcal{S} \cap B_R^c} 
[I(s)+\Phi_0(s)] \to +\infty$ as $R \to +\infty$, which means that 
 for $R$ big enough 
$\inf_{s \in O \cap B_R}
[I(s)+\Phi_0(s)-I_0]=\inf_{s \in O }
[I(s)+\Phi_0(s)-I_0]$.
Combining these facts the thesis follows. 
\end{proof}

\begin{proof}
 {\bf of  Proposition~\ref{LDP_postVara} \,\,} The proof follows by Lemma
~\ref{Lem_Varadhan}. To check the assumptions,
 note that 
 \[
 Q_1,\dots,Q_L \mapsto \Sigma(Q_1,\dots,Q_L)
=\begin{pmatrix}
\Sigma_{00} &  \Sigma_{01} \\
\Sigma_{01}^\top &  \Sigma_{11} \\
\end{pmatrix}
:=\frac{\tilde \bX^\top \tilde \bX}{N_0 \lambda^*} \otimes  Q^{(L)}  
 \]
 is a continuous function. 
Moreover, 
$\Sigma_{11} \mapsto 
\by_{1:P}^\top ( \Sigma_{11} +\beta^{-1}  \eI_{DP})^{-1} \by_{1:P}$
is continuous and bounded (on 
$\CS^+$) and positive, and hence the same holds for 
\[
Q_1,\dots,Q_L \mapsto
\by_{1:P}^\top ( \Sigma_{11}(Q_1,\dots,Q_L) +\beta^{-1}  \eI_{DP})^{-1} \by_{1:P}.
\]
Let $\lambda_i$'s be the eigenvalues of
$\beta \Sigma_{11}$, then $\eI_{DP}+\beta \Sigma_{11}$ has eigenvalues $1+\lambda_i$
and 
\[
\log(\det(\eI_{DP} +\beta \Sigma_{11})) =
\sum_{i=1}^{DP} \log(1+\lambda_i).
\]
hence 
$\Sigma_{11} \mapsto 
\log(\det(\eI_{DP} +\beta \Sigma_{11}))$
is  bounded on bounded  set of $\Sigma_{11}$.
Using the continuity of  $(Q_1,\dots,Q_L) \mapsto \Sigma_{11}(Q_1,\dots,Q_L)$
it follows that also $(Q_1,\dots,Q_L) \mapsto 
\log(\det(\eI_{DP} +\beta \Sigma_{11}))$
is bounded when $\sum_{\ell=1}^L 
\|Q_\ell\|_2 \leq R$ for any $R$. 
It remains to check that 
\[
\lim_{R \to +\infty} \inf_{(Q_1,\dots,Q_L): \sum
\|Q_\ell\|_2^2 > R^ 2 } \sum_{\ell=1}^L \Big(  \tr(Q_\ell)-
\log(\det(Q_\ell ))\Big)
 = +\infty.
 \]
Letting $\lambda_{i\ell}$ the eigenvalues
of $Q_\ell$, the claim follows since 
$\tr(Q_\ell)-\log(\det(Q_\ell ))=\sum_{i=1}^P\Big ( \lambda_{i\ell}+
\log(\lambda_{i\ell})\Big)$.
\end{proof}

\subsection{Proofs for the C-DLN}

\begin{lemma}\label{CNN_striclty_positive}
Let $\CNN(Q_1,\dots,Q_L)$ 
be defined as in \eqref{Def_CNN_matrix}. 
Then, if $Q_1,\dots,Q_L$
are strictly 
positive  matrices, 
then  $\CNN(Q_1,\dots,Q_L)$
is positive. 
If in addition  
$\sum_{a_0}\bX_{a_0}^\top \bX_{a_0}$ is  strictly 
positive, then 
$\CNN(Q_1,\dots,Q_L)$ is strictly 
positive. 
\end{lemma}

\begin{proof}
First of all note that 
if $Q_1,\dots,Q_L$ are strictly
positive then $\mathcal{T}(Q_1,\dots,Q_L)$ is strictly positive. 
For $\bar \bs_{1:P} \not=0$ 
write 
\[
\begin{split}
\bar \bs_{1:P}^\top \CNN(Q_1,\dots,Q_L)
\bar \bs_{1:P} 
& =
\sum_{\mu,\nu}
\sum_{a_0}
\sum_{r,s}\mathcal{T}(Q_1,\dots,Q_L)_{rs}
\frac{x^{\mu}_{a_0,r} x^{\nu}_{a_0,s}}
{\lambda^* C_0} \bar s_{\mu} \bar s_{\nu} 
\\
& = 
\sum_{a_0}
\sum_{r,s}\mathcal{T}(Q_1,\dots,Q_L)_{rs}
A^*_{r,a_0} A^*_{s,a_0} 
\end{split}
\]
with $A^*_{r,a_0}= 
\sum_{\mu} \bar s_{\mu} 
\frac{x^{\mu}_{a_0,r}}{\sqrt{\lambda^* C_0}}$.
Using  $\mathcal{T}(Q_1,\dots,Q_L)>0$,
one gets  
\[
\sum_{r,s}\mathcal{T}(Q_1,\dots,Q_L)_{rs}
A^*_{r,a_0} A^*_{s,a_0} \geq 0
\]
for every  $a_0$, so that   
$\bar \bs_{1:P}^\top \CNN(Q_1,\dots,Q_L)
\bar \bs_{1:P} \geq 0$. 
If now  
$0=\bar \bs_{1:P}^\top \CNN(Q_1,\dots,Q_L)
\bar \bs_{1:P}$,
then 
$A^*_{r,a_0}=0$ for every $r$ 
and $a_0$. Hence, 
\[
0=\sum_r (A^*_{r,a_0})^2=\sum_{\mu,\nu} \bar s_{\mu}  \bar s_{\nu}
\sum_{r} \frac{x^{\mu}_{a_0,r}x^{\nu}_{a_0,r}}{{\lambda^* C_0}}
\]
If 
$\sum_{a_0}\bX_{a_0}^\top \bX_{a_0}>0$, 
this is possible only 
if $\bar \bs_{1:P} =0$ . 
\end{proof}

Let $K=[K_{i,j,\mu,\nu}]_{i,j,\mu,\nu}$
a four index tensor with $i,j=1,\dots,N_0$ 
and $\mu,\nu=1,\dots,P$ corresponding 
to a covariance operator. 
One can identifies $K$ with a
symmetric and positive definite $(N_0\times P) 
\times (N_0 \times P) $ matrix 
$K=[K_{i,j,\mu,\nu}]_{(i,\mu),(j,\nu)}$, where the multi-indices $(i,\mu)$'s are properly ordered from $1$ to $N_0 \times P$. With a slight abuse of language, we will use in the following $K^{(\ell -1)}$ 
to denote either the four indices tensor or its matrix representation, the case being clear from the context.

\begin{lemma}\label{lemma_determinati}
Let $\mathbf{s}$ be a vector in $\RE^P$, then 
\begin{equation}
\begin{aligned}
    \det[\eI_{N_0}\otimes\eI_P + [\eI_{N_0}\otimes(\mathbf{s}\mathbf{s}^\top)] K] 
    &=\det \Big ( \eI_{N_0} + \sum_{\mu,\nu} {s}_\mu
    {s}_\nu K_{\cdot,\mu\nu} \Big)\,,
\end{aligned}
\end{equation}
where $K_{\cdot,\mu\nu} = [K_{i,j,\mu,\nu}]_{i,j=1}^{N_0}$.
\end{lemma}

\begin{proof}
The eigenvalues of $[\eI_{N_0} \otimes (\mathbf{s}\mathbf{s}^\top)] K$
are given by 
\[
[\eI_{N_0} \otimes (\mathbf{s}\mathbf{s}^\top)] K v =\lambda v
\]
which reads, in components, as
\[
\sum_{\mu,\nu,j} s_{\mu} K_{(i,\mu)(j,\nu)} v_{(j,\nu)} s_{\rho}  =\lambda v_{(i,\rho)}\,.
\]
This shows that if $\lambda\not=0$ then 
$v_{(i,\mu)} =c_i s_\mu$ for some $\mathbf{c} \in \mathbb{R}^{N_0}$, such that
\[
\sum_j \Biggl( \sum_{\mu,\nu} s_{\mu} K_{\cdot,\mu\nu} s_{\nu}\Biggr)_{ij} c_{j}    =\lambda c_{i}\,,
\]
that is
$\lambda \in \Sp(\sum_{\mu,\nu} s_\mu K_{\cdot,\mu\nu} s_\nu)$, where $\Sp$ denotes the spectrum. The cardinality of this spectrum is $N_0$, all the other $N_0(P-1)$ eigenvalues must be zero. Hence,
\[
\Sp(\eI_{N_0}\otimes \eI_P + [\eI_{N_0} \otimes (\mathbf{s}\mathbf{s}^\top)] K) = \Sp\Bigl(\eI_{N_0}+\sum_{\mu,\nu} s_\mu K_{\cdot,\mu\nu} s_{\nu}\Bigr) \cup \{1\}^{N_0(P-1)}\,,
\]
and the thesis follows. 
\end{proof}

\begin{proof}
{\bf of Proposition~\ref{prop1-CNN}}
From Eq.~\eqref{eq:CNN_S}, we have
\begin{equation}\label{starting_CNN}
    \E[e^{i \bar{ \mathbf{s}}_{1:P}^\top \mathbf{S}_{1:P}}] = \E \left[ \exp\left(-\frac{1}{2}
    \sum_{\mu,\nu} \bar{s}^\mu\bar{s}^\nu    \sum_{a_L=1}^{C_L} \sum_{i=1}^{N_0} \frac{1}{\lambda_L C_L N_0}    h^{(L)}_{a_L,i,\mu}h^{(L)}_{a_L,i,\nu}\right) \right]\,.
\end{equation}
 Let  $h^{(\ell)}$ be the collection of 
all the variables 
$h_{a_\ell,i,\mu}^{(\ell)}=h_{a_\ell,i}^{(\ell)}(\bx^{\mu})$
where $a_\ell=1,\dots,C_\ell$, $i=1,\dots,N_0$, $\mu=1,\dots,P$. 
By \eqref{convolutiona.recur1}, it is clear 
that, conditionally on $h^{(\ell-1)}$, 
 $h^{(\ell)}$ is a collection 
 of jointly Gaussian random variables 
 (a Gaussian field) with zero mean and covariance function given 
by  
\[
\Cov(h_{a,i,\mu},h_{b,j,\nu}|h^{(\ell-1)})=
\delta_{a,b} K_{i,j,\mu,\nu}^{(\ell -1)}
\]
where 
\begin{equation}
    K_{i,j,\mu,\nu}^{(\ell -1)} =  \frac{1}{ \lambda_{\ell-1} C_{\ell - 1} M}  \sum_{a_{\ell - 1} = 1}^{C_{\ell - 1}} \sum_{m = -\lfloor M/2 \rfloor}^{\lfloor M/2 \rfloor}\left( \sum_{r=1}^{N_0}T_{m,ir} h^{(\ell - 1)}_{a_{\ell-1},r,\mu} \right) \left( \sum_{s=1}^{N_0} T_{m,js} h^{(\ell - 1)}_{a_{\ell-1},s,\nu} \right).
\end{equation}
Considering $h^{(\ell)}$ as a vector 
of dimension $C_\ell \times P \times N_0$, 
 the conditional distribution 
 of  $h^{(\ell)}$ given  $h^{(\ell-1)}$
is 
\[
h^{(\ell)} | h^{(\ell-1)} \sim \mathcal{N}(0,\eI_{C_\ell}\otimes K^{(\ell-1)})
\]
(see the comment before Lemma~\ref{lemma_determinati} for the matrix representation of the tensor $K^{(\ell - 1)}$).
At this stage, by Gaussian integration, 
form \eqref{starting_CNN} one gets
\begin{equation}
\begin{aligned}
\label{eq:CNN_proof_dets}
    \E[e^{i \bar{ \mathbf{s}}_{1:P}^\top \mathbf{S}_{1:P}}] 
    &= \E \Biggl[ \det\left((\eI_{C_L}\otimes K^{(L-1)})^{-1}+\frac{1}{\lambda_L C_L N_0}\eI_{C_L}\otimes \eI_{N_0}\otimes (\bar{ \mathbf{s}}_{1:P} \bar{ \mathbf{s}}_{1:P}^\top ) \right)^{-\frac{1}{2}} \\
    & \qquad \qquad \qquad \qquad \det(\eI_{C_L}\otimes K^{(L-1)})^{-\frac{1}{2}}\Biggr]\\
    &= \E \Biggl[ \det\left(\eI_{N_0}\otimes \eI_{P}+\frac{1}{\lambda_L C_L N_0} [\eI_{N_0}\otimes (\bar{ \mathbf{s}}_{1:P} \bar{ \mathbf{s}}_{1:P}^\top)]K^{(L-1)} \right)^{-\frac{C_L}{2}} \Biggr]\\
    &=\E \Biggl[ \det\Biggl(  \eI_{N_0} + \frac{1}{\lambda_L C_L N_0} \sum_{\mu,\nu} \bar{ s}_\mu \bar{ s}_{\nu} K^{(L-1)}_{\cdot,\mu\nu} \Biggr)^{-\frac{C_L}{2}} \Biggr]\,.
\end{aligned}
\end{equation}
The last step follows from 
Lemma~\ref{lemma_determinati}.
Using Eq.~\eqref{LaplaceWishart}, we get
\begin{equation}
    \begin{aligned}
       &  \E[e^{i \bar{ \mathbf{s}}_{1:P}^\top \mathbf{S}_{1:P}}]  \\
       & 
    \quad = \E \Biggl[ \int_{\CS^+_{N_0}} \exp\Biggl\{- \frac{1}{2\lambda_L N_0} \sum_{i,j}Q_{L,ij}  \sum_{\mu,\nu} \bar{ s}_\mu \bar{ s}_{\nu} K^{(L-1)}_{ij,\mu\nu}\Biggr\} \CW_{N_0}\Big (dQ_L|\frac{1}{C_L}\eI_{N_0},C_L \Big)\Biggr ]\\
    &\quad = \E \Biggl[ \int_{\CS^+_{N_0}} \exp\Biggl\{- \frac{1}{2\lambda_L\lambda_{L-1} C_{L - 1} N_0} \sum_{i,j} 
    Q^*_{L,ij}
    \sum_{\mu,\nu} \bar{ s}_\mu \bar{ s}_{\nu}   \sum_{a_{L - 1} = 1}^{C_{L - 1}}  h^{(L - 1)}_{a_{L-1},i,\mu}  h^{(L - 1)}_{a_{L-1},j,\nu}\Biggr\}  \\
    & \qquad \qquad \qquad \qquad \qquad\CW_{N_0}\Big(dQ_L|\frac{1}{C_L}\eI_{N_0},C_L\Big)
    \Biggr ]
    \end{aligned}
\end{equation}
where we used Eq.~\eqref{eq:CNN_Qstar}.

We could in principle proceed directly integrating out $h^{(L-1)}$ as we did for $h^{(L)}$, Eq.~\eqref{eq:CNN_proof_dets}; however, in this way we would end up with a Wishart measure for $Q_{L-1}$ dependent on $Q_L$. To keep factorized the different contributions, it is easier to change variables with the linear transformation
\begin{equation}
    \tilde{h}^{(L-1)}_{a_{L-1},i,\mu} = \sum_j U^*_{L,ij} h^{(L-1)}_{a_{L-1},j,\mu}\,,
\end{equation}
where $(U^*_{L})^\top U^*_{L} = Q^*_L $. It is clear that
the conditional distribution of 
$\tilde{h}^{(L-1)}$ given $h^{(L-2)}$
is again Gaussian, more precisely 
\begin{equation}
    \tilde{h}^{(L-1)}|h^{(L-2)}  
    \sim \mathcal{N}(0,\eI_{C_{L-1}}\otimes[U^*_{L} K^{(L-2)}(U^*_{L})^\top ] )\,,
\end{equation}
with the slight abuse of notation $U^*_{L} \equiv U^*_{L} \otimes \eI_P$. Proceeding as before,
\begin{equation}
    \begin{aligned}
    \E[e^{i \bar{ \mathbf{s}}_{1:P}^\top \mathbf{S}_{1:P}}] 
    &= \E \Biggl[ \int_{(\CS^+_{N_0})^2} \exp\Biggl
    \{- \frac{1}{2\lambda_L\lambda_{L-1} N_0} \tr\Biggl( Q_{L-1}  \sum_{\mu,\nu} \bar{ s}_\mu \bar{ s}_{\nu} [U^*_L K^{(L-2)}_{\cdot,\mu\nu}(U^*_L)^\top ]\Biggr) \Biggr\}  \\
    &\qquad \qquad  \prod_{\ell \in \{ L, L-1 \}}\CW_{N_0}(dQ_{\ell}|\frac{1}{C_{\ell}}\eI_{N_0},C_{\ell}) \Biggr]
    \\
    &= \E \Biggl[  \int_{(\CS^+_{N_0})^2} \exp\Biggl\{- \frac{1}{2\lambda_L\lambda_{L-1}\lambda_{L-2} C_{L - 2} N_0} \sum_{i,j} 
    Q^*_{L-1,ij}
    \sum_{\mu,\nu} \bar{ s}_\mu \bar{ s}_{\nu}   \\
    & \qquad \qquad \sum_{a_{L - 2} = 1}^{C_{L - 2}}  h^{(L - 2)}_{a_{L-2},i,\mu}  h^{(L - 2)}_{a_{L-2},j,\nu}\Biggr\} 
    \times\prod_{\ell \in \{ L, L-1 \}}\CW_{N_0}\Big (dQ_{\ell}|\frac{1}{C_{\ell}}\eI_{N_0},C_{\ell}\Big)\Biggr ]
    \end{aligned}
\end{equation}
where $Q^*_{L-1}$ is again given by Eq.~\eqref{eq:CNN_Qstar}. Proposition~\ref{prop1-CNN} is proven once this procedure is straightforwardly iterated.
\end{proof}

\begin{proof}{\bf of Proposition~\ref{Prop2-CNN}}
The proof  of
Proposition~\ref{Prop2-CNN}
is identical to the proof of 
Proposition~\ref{Prop2}
once one observes that 
the matrix $\Sigma$,
which is $\CNN(Q_1,\dots,Q_L)$
for the enlarged dataset 
$\tilde \bX$, is strictly positive 
by Lemma~\ref{CNN_striclty_positive}.\end{proof}


\vskip 0.2in
\bibliography{biblio_JMLR}

\end{document}